\documentclass{article}

\usepackage{microtype}
\usepackage{graphicx}
\usepackage{subfigure}
\usepackage{booktabs} % for professional tables
\usepackage{amsmath}
\usepackage{amssymb} 
\usepackage{cite}
\usepackage{amsthm} 
\usepackage{algorithm}
\usepackage{algorithmic}
\usepackage{multirow}
\usepackage{xcolor}
\usepackage{colortbl}
\usepackage{xspace}
\newcommand{\ourtool}{\textsc{VeriMoA}\xspace}
\usepackage{enumitem}

\usepackage{hyperref}

\usepackage[accepted]{mlsys2025}

\mlsystitlerunning{Submission for MLSys 2026}

\begin{document}

\twocolumn[
\mlsystitle{\ourtool:\\A Mixture-of-Agents Framework for Spec-to-HDL Generation}

% It is OKAY to include author information, even for blind
% submissions: the style file will automatically remove it for you
% unless you've provided the [accepted] option to the mlsys2025
% package.

% List of affiliations: The first argument should be a (short)
% identifier you will use later to specify author affiliations
% Academic affiliations should list Department, University, City, Region, Country
% Industry affiliations should list Company, City, Region, Country

% You can specify symbols, otherwise they are numbered in order.
% Ideally, you should not use this facility. Affiliations will be numbered
% in order of appearance and this is the preferred way.
% \mlsyssetsymbol{equal}{*}

\begin{mlsysauthorlist}
\mlsysauthor{Heng Ping}{usc}
\mlsysauthor{Arijit Bhattacharjee}{isu}
\mlsysauthor{Peiyu Zhang}{usc}
\mlsysauthor{Shixuan Li}{usc}
\mlsysauthor{Wei Yang}{usc}
\mlsysauthor{Anzhe Cheng}{usc}
\mlsysauthor{Xiaole Zhang}{usc}
\mlsysauthor{Jesse Thomason}{usc}
\mlsysauthor{Ali Jannesari}{isu}
\mlsysauthor{Nesreen Ahmed}{cisco}
\mlsysauthor{Paul Bogdan}{usc}
\end{mlsysauthorlist}

\mlsysaffiliation{usc}{University of Southern California, Los Angeles, CA, USA}
\mlsysaffiliation{isu}{Iowa State University, Ames, IA, USA}
\mlsysaffiliation{cisco}{Cisco AI Research, USA}

\mlsyscorrespondingauthor{Paul Bogdan}{pbogdan@usc.edu}

% You may provide any keywords that you
% find helpful for describing your paper; these are used to populate
% the "keywords" metadata in the PDF but will not be shown in the document
\mlsyskeywords{Machine Learning, MLSys}

\vskip 0.3in

\begin{abstract}
Automation of Register Transfer Level (RTL) design can help developers meet increasing computational demands. Large Language Models (LLMs) show promise for Hardware Description Language (HDL) generation, but face challenges due to limited parametric knowledge and domain-specific constraints. While prompt engineering and fine-tuning have limitations in knowledge coverage and training costs, multi-agent architectures offer a training-free paradigm to enhance reasoning through collaborative generation. However, current multi-agent approaches suffer from two critical deficiencies: susceptibility to noise propagation and constrained reasoning space exploration. We propose \textbf{\ourtool}, a training-free mixture-of-agents (MoA) framework with two synergistic innovations. First, a \textbf{quality-guided caching mechanism} to maintain all intermediate HDL outputs and enables quality-based ranking and selection across the entire generation process, encouraging knowledge accumulation over layers of reasoning. Second, a \textbf{multi-path generation strategy} that leverages C++ and Python as intermediate representations, decomposing specification-to-HDL translation into two-stage processes that exploit LLM fluency in high-resource languages while promoting solution diversity. Comprehensive experiments on VerilogEval 2.0 and RTLLM 2.0 benchmarks demonstrate that \ourtool achieves 15--30\% improvements in Pass@1 across diverse LLM backbones, especially enabling smaller models to match larger models and fine-tuned alternatives without requiring costly training.
\end{abstract}
]

% this must go after the closing bracket ] following \twocolumn[ ...

% This command actually creates the footnote in the first column
% listing the affiliations and the copyright notice.
% The command takes one argument, which is text to display at the start of the footnote.
% The \mlsysEqualContribution command is standard text for equal contribution.
% Remove it (just {}) if you do not need this facility.

%\printAffiliationsAndNotice{}  % leave blank if no need to mention equal contribution
\printAffiliationsAndNotice{} % otherwise use the standard text.
\section{Introduction}

As the semiconductor industry faces unprecedented pressure to deliver high-performance hardware, the automation of Register Transfer Level (RTL) design, a critical component of the chip development pipeline, has become increasingly vital~\cite{liu2023chipnemo}. The ability to automatically generate accurate Hardware Description Language (HDL) from natural language specifications would dramatically accelerate design cycles and reduce engineering pressure~\cite{lu2024rtllm}. Recent advances in Large Language Models (LLMs) have demonstrated remarkable capabilities across diverse domains~\cite{kanakaris2025network,yang2025toward,duan2023leveraging}, including Electronic Design Automation (EDA), where they have shown promising potential for automating HDL generation from natural language instructions~\cite{ thakur2023autochip}.
Unfortunately, despite their impressive performance in general-purpose programming tasks, applying LLMs to HDL generation presents unique challenges. 

Unlike widely-used languages such as C++ or Python, HDL represents a specialized domain with limited representation in LLM pretraining corpora, resulting in comparatively sparse parametric knowledge for hardware design~\cite{liu2024openllm}. Moreover, HDL exhibits distinct operational semantics and constraints.
For example, HDL must capture concurrent hardware behavior, satisfy timing constraints, and adhere to synthesis requirements, requiring reasoning over temporal and physical limitations inherent to circuit design. These domain-specific characteristics make it difficult for general-purpose LLMs to generate correct and efficient HDL code from specifications~\cite{alsaqer2024potential}.

To address these limitations, initial research focused on \textit{model-centric} strategies to enhance LLM's HDL generation capabilities. The first wave employed prompt engineering to elicit the LLM's sparse HDL knowledge: ParaHDL~\cite{sun2025paradigm} proposed task-specific prompt paradigms embedding HDL syntax patterns, while AoT~\cite{delorenzo2025aot} used tailored Chain-of-Thought templates for different circuit categories. However, these methods are fundamentally limited by the LLM's inadequate pre-existing HDL understanding. A second wave progressed towards augmenting parametric knowledge through fine-tuning on high-quality HDL datasets. Methods like RTLCoder~\cite{liu2024rtlcoder} and AutoVCoder~\cite{gao2024autovcoder} curated large-scale, simulation-verified HDL corpora for supervised fine-tuning, while VeriSeek~\cite{wang2024veriseek} and ChipSeek-R1~\cite{chen2025chipseek} employed reinforcement learning to further enhance reasoning. Despite substantial improvements, the model-centric philosophy faces critical limitations: prompt engineering remains bounded by limited pre-existing knowledge; fine-tuning requires significant data curation and training costs; most fundamentally, both rely on monolithic generation, lacking mechanisms for systematic collaboration and progressive improvement.

Recognizing the risks of a monolithic generation process, an alternative research direction constructs system-level frameworks, with multi-agent architectures emerging as a prominent approach for HDL generation. However, the prevailing collaborative paradigms are fundamentally flawed, ranging from the rigid, linear pipelines of MAGE~\cite{zhao2024mage} that propagate errors, to the unstructured debates of CoopetitiveV~\cite{mi2024coopetitivev} that lead to chaotic exploration. This progression exposes two profound deficiencies in current approaches. First, the collaborative processes of these frameworks are highly susceptible to noise and cannot guarantee 
information gain. Without a principled mechanism to filter out flawed information, 
the system's knowledge base can be easily corrupted, preventing it from productively 
building upon prior successes. Second, even when correct information is preserved, 
they operate within a constrained reasoning space. Both brittle and 
chaotic structures struggle to systematically navigate the vast design space of HDL, 
a limitation that causes them to converge prematurely on local optima. The failure 
of current paradigms to both effectively capitalize on valuable findings and 
adequately explore the solution space highlights the critical need for a more 
advanced framework.

To overcome these fundamental deficiencies, we propose \textbf{\ourtool (Quality-guided Multi-path Mixture-of-Agents for HDL Generation)}, a framework that systematically addresses both noise susceptibility and constrained reasoning space in current multi-agent approaches. Building upon the Mixture-of-Agents (MoA) paradigm~\cite{wang2024mixture}, which aggregates outputs from the preceding layer, we introduce two key innovations. While MoA's layer-wise aggregation partially mitigates error propagation compared to linear pipelines, it still suffers from cascaded dependencies where hallucination errors accumulate across layers. We address noise susceptibility through a \textbf{quality-guided caching mechanism} that fundamentally breaks cascaded dependencies by evaluating and caching intermediate HDL outputs from all previous layers, enabling deeper agents to selectively leverage the highest-quality code from the entire process rather than being confined to the immediately preceding layer. For constrained reasoning space, we introduce \textbf{multi-path generation strategies} within each layer, incorporating intermediate representation paths via C++ and Python that decompose specification-to-HDL translation into two-stage processes (specification $\rightarrow$ high-level code $\rightarrow$ HDL), substantially expanding the solution space at each layer.

These two innovations synergistically enhance HDL generation performance. The quality-guided caching ensures monotonic knowledge accumulation across layers, preventing valuable insights from being lost. The multi-path generation leverages LLMs' rich parametric knowledge in C++ and Python while introducing structured diversity across generation paths (direct HDL, C++-guided, Python-guided) for comprehensive solution space exploration. Additionally, our framework can integrate simulation-based self-refinement to further enhance code quality, culminating in superior outputs. 

Our contributions are summarized as follows:

\begin{itemize}[nosep,noitemsep]
    \item \textbf{Quality-Guided Mixture-of-Agents Architecture}: We propose a novel multi-agent framework that combines quality-guided caching with layered progress, ensuring monotonic knowledge accumulation and preventing information loss across layers.
    \item \textbf{Multi-Path Generation with Intermediate Representations}: We introduce parallel generation paths incorporating C++ and Python as intermediate representations, enabling two-stage HDL synthesis that leverages LLMs' strengths in high-level languages while promoting solution diversity.
    \item \textbf{Comprehensive Empirical Validation}: Through experiments on VerilogEval 2.0 and RTLLM 2.0 benchmarks, we demonstrate \ourtool achieves superior performance with 15--30\% improvements in Pass@1 across diverse LLM backbones, validating the effectiveness of our quality-guided multi-path approach.
\end{itemize}
\section{Related Work}
\subsection{LLMs for HDL Generation}
Existing LLM-based HDL generation approaches can be categorized into three primary strategies: prompt engineering, fine-tuning, and multi-agent architectures.

\textbf{Prompt Engineering.} Given the limited parametric knowledge of HDL in general-purpose LLMs, prompt engineering approaches inject HDL programming guidelines directly into prompts. ParaHDL~\cite{sun2025paradigm} and ClassHDL~\cite{sun2024classhdl} developed distinct prompt paradigms for combinational and sequential logic circuits. %HDLCoRe~\cite{ping2025hdlcore} extended this by incorporating difficulty-adaptive prompts and a two-stage heterogeneous RAG system. 
AoT~\cite{delorenzo2025aot} further refined this by subdividing combinational logic into fine-grained categories (Truth Table, Boolean, K-map, Multiplexers) with specialized paradigms for each, while providing detailed FSM paradigms for sequential logic. While lightweight and effective, prompt engineering remains fundamentally bounded by LLMs' limited intrinsic HDL understanding.

\textbf{Fine-Tuning Methods.} Fine-tuning approaches enhance LLMs' parametric HDL knowledge through dataset construction and advanced training methodologies. For datasets, MG-Verilog~\cite{zhang2024mgverilog}, OriGen~\cite{cui2024origen}, CodeV~\cite{zhao2024codev}, RTLCoder~\cite{liu2024rtlcoder}, AutoVCoder~\cite{gao2024autovcoder}, HaVen~\cite{yang2025haven}, CraftRTL~\cite{deng2025scalertl}, and PyraNet~\cite{nadimi2024pyranet} leverage existing HDL datasets, augmenting them through LLM-based generation and simulator-based validation. ReasonV~\cite{qin2025reasoningv}, ScaleRTL~\cite{deng2025scalertl}, VeriThoughts~\cite{yubeaton2025verithoughts}, and VeriRL~\cite{teng2025verirl} further generate reasoning-annotated datasets using commercial LLMs. For training methodologies, ReasonV~\cite{qin2025reasoningv} proposes two-stage instruction fine-tuning, VeriSeek~\cite{wang2024veriseek} applies code-structure-guided reinforcement learning, and ChipSeek-R1~\cite{chen2025chipseek} employs hierarchical reinforcement learning. Despite their effectiveness, these methods require substantial computational costs.

\textbf{Multi-Agent Architectures.} PromptV~\cite{mi2024promptv}, MAGE~\cite{zhao2024mage}, AutoSilicon~\cite{li2025autosilicon}, PRO-V~\cite{zhao2025prov}, HDLCoRe~\cite{ping2025hdlcore}, VeriOpt~\cite{tasnia2025veriopt} and AutoVeriFix~\cite{tan2025autoverifix} decompose HDL generation into sequential stages with specialized agents for generation, testbench creation, and debugging. 
% However, this sequential paradigm suffers from error accumulation as hallucinations propagate linearly across agents. 
CoopetitiveV~\cite{mi2024coopetitivev} employs debate mechanisms but risks domination by stronger LLMs. VeriMaAS~\cite{bhattaram2025verimaas} adaptively selects agentic operators based on task complexity. While achieving competitive results without training overhead, these approaches face critical limitations: sequential dependencies cause error propagation, while unstructured debates risk suboptimal exploration.

\subsection{Mixture-of-Agents (MoA)} 
The Mixture-of-Agents paradigm~\cite{wang2024mixture} aggregates complete responses from multiple LLMs through prompt-based synthesis, eliminating fine-tuning requirements while enabling heterogeneous model integration. Unlike Mixture-of-Experts (MoE)~\cite{shazeer2017outrageously} which requires gated routing among expert sub-networks, MoA operates at the model output level. It employs a layered architecture where proposers generate diverse responses and an aggregator synthesizes them into final solutions. This structured feed-forward flow distinguishes MoA from ranker-based methods~\cite{jiang2023llm} that simply select the best response, and debate-based approaches~\cite{du2023improving,liang2023encouraging} that rely on interactive discussions.

Recent analysis~\cite{li2025rethinking} reveals that MoA performance correlates strongly with both quality and diversity of the proposers, with quality showing higher correlation. This has led to two variants of standard MoA: Mixed-MoA aggregates outputs from diverse heterogeneous models, while Self-MoA samples multiple outputs from a single high-quality model. Both achieve state-of-the-art performance on instruction-following and dialogue benchmarks. However, standard MoA faces a critical limitation: unstable information propagation across layers, where valuable solutions can be lost due to hallucination or weak aggregation.

\section{Problem Definition}

\begin{figure*}[t]
  \centering
  \includegraphics[width=\textwidth]{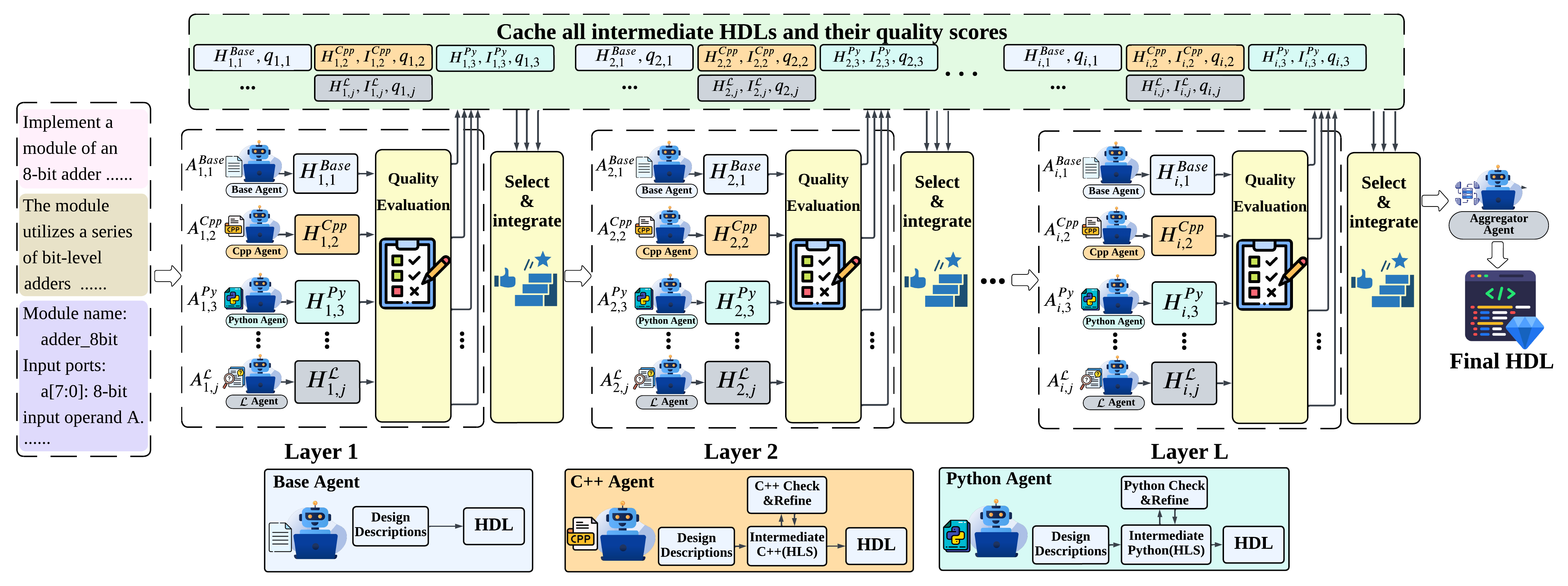}
  \caption{Overview of \ourtool. The framework employs L layers with M agents per layer, featuring three agent types (Base, C++, Python) that generate HDL through different paths. A global cache stores all intermediate outputs with quality scores, enabling quality-guided selection across layers. Agents at each layer receive top-n quality HDL codes from all previous layers, ensuring monotonic improvement. The final aggregator synthesizes high-quality candidates into the output.}
  \label{fig:main_framework}
\end{figure*}

We formalize the HDL generation task and establish the optimization objective for benchmark evaluation. Given a hardware design benchmark $\mathcal{B}$ consisting of $N$ design problems, each problem $i \in \{1, 2, \ldots, N\}$ is characterized by a tuple $(\mathcal{D}_i, \mathcal{T}_i, \mathcal{S}_i)$, where $\mathcal{D}_i$ denotes the natural language design description, $\mathcal{T}_i$ represents the golden testbench for simulation-based verification, and $\mathcal{S}_i$ denotes the simulator that validates functional correctness. The objective is to construct an LLM-based generation system $\mathcal{G}$ that produces HDL code $H_i$ for each design description $\mathcal{D}_i$:
\begin{equation}
H_i = \mathcal{G}(\mathcal{D}_i; \theta)
\end{equation}
where $\theta$ represents the system setting and generation strategy. The generated code $H_i$ is evaluated through simulation $\text{Pass}_i = \mathcal{S}_i(H_i, \mathcal{T}_i) \in \{\text{True}, \text{False}\}$, where $\text{Pass}_i = \text{True}$ indicates that $H_i$ passes all test cases, demonstrating both syntactic validity and functional correctness.

For practical deployment, the system generates multiple trials for each problem, and we measure performance using standard code generation metrics. Let $\mathcal{M}_k(\mathcal{G}, \mathcal{B})$ denote the pass@k metric that evaluates whether at least one correct solution exists among $k$ generated trials, averaged across all problems in benchmark $\mathcal{B}$. Our goal is to design a generation system $\mathcal{G}^*$ that maximizes this metric:
\begin{equation}
\label{eq:metric}
\mathcal{G}^* = \arg\max_{\mathcal{G}} \mathcal{M}_k(\mathcal{G}, \mathcal{B})
\end{equation}

\section{Quality-guided Multi-path Mixture-of-Agents for HDL Generation}

Existing multi-agent frameworks for HDL generation suffer from two critical deficiencies: susceptibility to noise propagation and constrained reasoning space exploration. To address these limitations, we propose \textbf{Quality-guided Multi-path Mixture-of-Agents  for HDL Generation (\ourtool)}, a framework that introduces two synergistic innovations: (1) a quality-guided caching mechanism ensuring monotonic knowledge accumulation across layers, and (2) multi-path generation strategies leveraging high-level code as intermediate representations for expanded solution space exploration. This section presents the implementation details of our framework, organized into two subsections corresponding to these core innovations.

\subsection{Quality-Guided Architecture with Global Caching}
\label{sec:quality_moa}

As shown in Fig.~\ref{fig:main_framework}, our quality-guided MoA consists of three interconnected components: MoA layers, a quality evaluator, and a global cache. The architecture is designed to break the cascaded dependencies of standard MoA while preserving structured information flow across layers.

\textbf{MoA Layers.} The framework employs a hierarchical structure comprising $L$ layers, organized into proposer layers (layers $1$ through $L-1$) and an aggregator layer (layer $L$). Each proposer layer contains $M$ agents operating in parallel to generate diverse HDL candidates. For an agent $A_{i,j}$ in the $i$-th proposer layer ($i \in \{1, \ldots, L-1\}$, $j \in \{1, \ldots, M\}$), the generation process is formalized as:
\begin{equation}
\label{eq:hdl_generation}
H_{i,j} = A_{i,j}(\mathcal{P}_{i,j})
\end{equation}
where $\mathcal{P}_{i,j}$ denotes the prompt and $H_{i,j}$ represents the HDL output. Following standard MoA conventions~\cite{wang2024mixture}, we maintain consistent agent configurations across proposer layers ($A_{i,j} = A_{i',j}$ for all $i, i' \in \{1, \ldots, L-1\}$). The aggregator synthesizes high-quality candidates from previous layers into the final HDL output.

\textbf{Quality Evaluation.} The quality evaluator $\mathcal{Q}$ assesses each generated HDL candidate through simulation-based evaluation, producing a quality score $q_{i,j} = \mathcal{Q}(H_{i,j}, \mathcal{T}, \mathcal{S})$. This score reflects both syntactic validity and functional correctness, serving as the foundation for quality-guided selection. The evaluator implements a hierarchical scoring strategy detailed in Algorithm~\ref{alg:quality_eval}, which incorporates HDL domain-specific knowledge through three evaluation paths: (1) perfect scores for fully correct implementations, (2) severity-weighted penalties for functional failures with valid syntax, and (3) rule-based fallback evaluation for syntactic failures. The quality scores encode hardware design principles such as reset signal requirements, signal driving conflicts and timing constraints into explicit evaluation criteria, providing agents with HDL-specific feedback that injects domain expertise lacking in general-purpose LLMs.

\begin{algorithm}[t]
\caption{Quality Evaluation with HDL-Specific Criteria}
\label{alg:quality_eval}
\begin{algorithmic}[1]
\REQUIRE HDL code $H$, testbench $\mathcal{T}$, simulator $\mathcal{S}$
\ENSURE Quality score $q \in [0, q_{\text{perfect}}]$
\STATE $\text{pass}_{\text{syntax}} \gets \text{SyntaxTest}(H, \mathcal{S})$
\STATE $\text{pass}_{\text{func}} \gets \text{FunctionTest}(H, \mathcal{T}, \mathcal{S})$
\IF{$\text{pass}_{\text{syntax}} = \text{True}$ \textbf{and} $\text{pass}_{\text{func}} = \text{True}$}
    \STATE \textbf{return} $q = q_{\text{perfect}}$
\ELSIF{$\text{pass}_{\text{syntax}} = \text{True}$ \textbf{and} $\text{pass}_{\text{func}} = \text{False}$}
    \STATE $p_{\text{severe}} \gets \text{EvaluateSevereErrors}(H)$ \COMMENT{Logic errors, $\in [0, \alpha_{\text{severe}}]$}
    \STATE $p_{\text{moderate}} \gets \text{EvaluateSynthesisIssues}(H)$ \COMMENT{Synthesis concerns, $\in [0, \alpha_{\text{moderate}}]$}
    \STATE $p_{\text{minor}} \gets \text{EvaluateStyleIssues}(H)$ \COMMENT{Style issues, $\in [0, \alpha_{\text{minor}}]$}
    \STATE \textbf{return} $q = q_{\text{base}} - p_{\text{severe}} - p_{\text{moderate}} - p_{\text{minor}}$
\ELSE
    \STATE $q_{\text{structure}} \gets \text{CheckModuleStructure}(H)$ \COMMENT{Module/IO, $\in [0, \beta_{\text{structure}}]$}
    \STATE $q_{\text{logic}} \gets \text{CheckLogicKeywords}(H)$ \COMMENT{Logic constructs, $\in [0, \beta_{\text{logic}}]$}
    \STATE $q_{\text{format}} \gets \text{CheckFormatting}(H)$ \COMMENT{Formatting, $\in [0, \beta_{\text{format}}]$}
    \STATE \textbf{return} $q = q_{\text{structure}} + q_{\text{logic}} + q_{\text{format}}$
\ENDIF
\end{algorithmic}

\end{algorithm}

\textbf{Global Cache and Quality-Guided Selection.} To address the unstable information propagation in standard MoA, we propose a global cache that maintains all intermediate outputs and enables quality-based selection across the entire generation process. All intermediate HDL outputs and their corresponding quality scores are stored in a global cache:
\begin{equation}
\label{eq:cache}
\mathcal{C} = \{(H_{i,j}, q_{i,j}) \mid i \in \{1, \ldots, L-1\}, j \in \{1, \ldots, M\}\}
\end{equation}
This cache decouples information flow from rigid layer-by-layer dependencies, enabling agents to access high-quality references from any previous layer. For the first proposer layer, agents receive only the design description: $\mathcal{P}_{1,j} = \mathcal{D}$. For subsequent layers ($i \geq 2$), each agent receives an augmented prompt:
\begin{equation}
\label{eq:prompt_augment}
\mathcal{P}_{i,j} = \mathcal{D} \oplus \mathcal{H}_{i}^{(n)}
\end{equation}
where $\oplus$ denotes concatenation and $\mathcal{H}_{i}^{(n)} = \text{TopN}(\mathcal{C}_{<i}, n)$ represents the top-$n$ highest-quality HDL codes selected from all previous layers. The TopN operation is defined as:
\begin{equation}
\label{eq:topn_hdl}
\text{TopN}(\mathcal{C}_{<i}, n) = \{H \mid (H, q) \in \mathcal{C}_{<i}, q \in \text{top-}n \text{ scores}\}
\end{equation}
where $\mathcal{C}_{<i} = \{(H_{k,j}, q_{k,j}) \mid k < i, j \in \{1, \ldots, M\}\}$ denotes the cache subset containing outputs from all layers before layer $i$. This quality-guided selection ensures that deeper agents always access the best available references, thereby breaking cascaded dependencies and enabling monotonic knowledge accumulation.

\subsection{Multi-Path Generation with High-level Code as Intermediate Representations}
\label{sec:multipath}

While the quality-guided caching mechanism ensures monotonic knowledge accumulation at the architectural level, we further enhance the framework by improving proposer layers internally through the multi-path generation strategy. The strategy leverages high-level programming languages as intermediate representations, enabling LLMs to utilize their superior C++/Python knowledge while exploring diverse reasoning paths within each layer. Let $\mathcal{L} \in \{\text{C++}, \text{Python}\}$ denote the intermediate language. For an agent following the $\mathcal{L}$-path in layer $i$, generation proceeds in two stages:

\textbf{Stage 1: Specification to Intermediate Code.}\label{sec:stage1} 
To further leverage the quality-guided caching mechanism, we evaluate and cache intermediate codes analogously to HDL codes, storing them in intermediate code cache $\mathcal{C}_{\mathcal{L}} = \{(I_{i,j}^{\mathcal{L}}, q_{i,j}^{\mathcal{L}}) \mid i \in \{1, \ldots, L-1\}, j \in \{1, \ldots, M_{\mathcal{L}}\}\}$ for each language $\mathcal{L}$. The agent generates intermediate code as:
\begin{equation}
\label{eq:intermediate_gen}
I_{i,j}^{\mathcal{L}} = A_{i,j}^{\text{Stage1}}(\mathcal{P}_{i,j}^{\mathcal{L}})
\end{equation}
where the prompt is constructed as:
\begin{equation}
\label{eq:intermediate_prompt}
\mathcal{P}_{i,j}^{\mathcal{L}} = \begin{cases}
\mathcal{D} & \text{if } i = 1 \\
\mathcal{D} \oplus \mathcal{I}_{i}^{\mathcal{L},(k)} & \text{if } i \geq 2
\end{cases}
\end{equation}
For the first layer, the agent generates intermediate code directly from the design description. For subsequent layers ($i \geq 2$), the prompt is augmented with $\mathcal{I}_{i}^{\mathcal{L},(k)} = \text{TopK}(\mathcal{C}_{\mathcal{L},<i}, k)$, which represents the top-$k$ highest-quality intermediate codes selected from all previous layers. 
We then perform syntax validation using checking scripts and invoke the agent for self-refinement based on checking feedback, producing refined intermediate code $I_{i,j}^{\mathcal{L},\text{refined}}$. 
% This leverages LLMs' strong understanding of C++/Python to correct syntactic issues before HDL generation.

\textbf{Stage 2: Intermediate Code to HDL.}\label{sec:stage2}  
Using the refined intermediate code as a reference, the agent generates HDL:
\begin{equation}
\label{eq:hdl_from_intermediate}
H_{i,j}^{\mathcal{L}} = A_{i,j}^{\text{Stage2}}(\mathcal{D} \oplus I_{i,j}^{\mathcal{L},\text{refined}} \oplus \mathcal{H}_{i}^{(n)})
\end{equation}
where $\mathcal{H}_{i}^{(n)}$ denotes the top-$n$ HDL references from the global cache. For $i=1$, this simplifies to $H_{1,j}^{\mathcal{L}} = A_{1,j}^{\text{Stage2}}(\mathcal{D} \oplus I_{1,j}^{\mathcal{L},\text{refined}})$ as no prior HDL references exist.

To enable quality-guided selection of intermediate codes across layers, we assign each intermediate code a quality score based on the HDL it generates:
\begin{equation}
\label{eq:intermediate_quality}
q_{i,j}^{\mathcal{L}} = q(H_{i,j}^{\mathcal{L}})
\end{equation}
where $q(H_{i,j}^{\mathcal{L}})$ is evaluated by the evaluator $\mathcal{Q}$ described in Algorithm~\ref{alg:quality_eval}. This task-aligned evaluation prioritizes intermediate codes that enable high-quality HDL translation.

\textbf{Multi-Path Proposer Configuration.} We define three agents: \textbf{Baseline agents} ($A_{i,j}^{\text{Base}}$) generate HDL directly from the design description, \textbf{C++ agents} ($A_{i,j}^{\text{Cpp}}$) and \textbf{Python agents} ($A_{i,j}^{\text{Py}}$) follow the two-stage process with C++ and Python as intermediate languages, respectively.
Each proposer layer consists of $M$ agents operating in parallel, with each position $j \in \{1, \ldots, M\}$ assigned one agent type:
\begin{equation}
\label{eq:agent_mixture}
\mathcal{A}_i = \{A_{i,j}^{t_j} \mid j \in \{1, \ldots, M\}, t_j \in \{\text{Base}, \text{Cpp}, \text{Py}\}\}
\end{equation}
This heterogeneous mixture explores multiple reasoning paths simultaneously, where baseline agents prioritize hardware idioms, C++ agents emphasize bit-level control, while Python agents leverage high-level expressiveness, leading to diverse HDL implementations for identical specifications.

\textbf{Optional Simulator-Based Self-Refinement.} Following established practices~\cite{zhao2024mage,mi2024coopetitivev}, our framework can optionally integrate simulator-based self-refinement. After generating initial HDL output $H_{i,j}$, agents invoke simulator $\mathcal{S}$ with testbench $\mathcal{T}$ to obtain execution feedback for refinement. This mechanism applies uniformly to all agent types in both proposer and aggregator layers.
\section{Theoretical Analysis of Quality-Guided Multi-Path Generation}

This section provides theoretical analysis of our \ourtool framework, establishing formal guarantees for performance improvement and explaining how quality-guided caching and multi-path generation enhance HDL generation. We connect our design to established Mixture-of-Agents principles, prove monotonic quality guarantees, and analyze how multi-path strategies amplify both quality and diversity.

\subsection{Foundation: MoA Performance Decomposition}

Recent empirical analysis of Mixture-of-Agents~\cite{li2025rethinking} demonstrates that MoA performance exhibits strong positive correlation with two key factors: proposer quality and proposer diversity. Through statistical analysis across reasoning benchmarks, they establish that MoA performance $t$ can be approximated by a linear model:
\begin{equation}
\label{eq:moa_performance}
t = \alpha \times q + \beta \times d + \gamma
\end{equation}
where $q$ represents proposer quality, $d$ represents proposer diversity, and $\alpha, \beta, \gamma$ are regression coefficients with $\alpha > \beta$, indicating quality exerts stronger influence than diversity.

In HDL generation, proposer quality encompasses agent capability and reference code quality from previous layers. Let $\mathcal{A}_i = \{A_{i,1}, \ldots, A_{i,M}\}$ denote the agent configuration at layer $i$, and $\mathcal{H}_i^{(n)}$ the reference HDL set. The effective proposer quality is:
\begin{equation}
\label{eq:proposer_quality}
Q_i = f(\text{capability}(\mathcal{A}_i), \text{quality}(\mathcal{H}_i^{(n)}))
\end{equation}
where $f$ captures the combined effect of agent capability and reference quality. With fixed agent configurations across layers, improving $Q_i$ reduces to enhancing $\text{quality}(\mathcal{H}_i^{(n)})$. Given $\alpha > \beta$, our framework prioritizes maximizing reference quality through quality-guided caching while promoting diversity through multi-path generation.

\subsection{Quality Guarantee through Global Caching}

\textbf{Error Accumulation in Standard MoA.} Applying standard MoA to HDL generation causes performance degradation as layer depth increases. This occurs because LLMs' limited HDL knowledge causes each layer to introduce non-trivial errors. In standard MoA, each proposer layer only receives outputs from the preceding layer without quality-based selection. Consequently, low-quality outputs from layer $i-1$ contaminate layer $i$'s reference set, harming final performance through cascaded error propagation.

\textbf{Monotonic Quality Improvement.} Our quality-guided caching mechanism addresses error accumulation by maintaining a global cache and selecting only top-$n$ quality outputs across all previous layers, ensuring monotonic improvement while breaking rigid layer-by-layer propagation.

For any layer $i \geq 2$, the quality guarantee can be formalized as:
\begin{equation}
\label{eq:min_quality}
\min_{H \in \mathcal{H}_{i+1}^{(n)}} q(H) \geq \min_{H \in \mathcal{H}_{i}^{(n)}} q(H)
\end{equation}
\begin{equation}
\label{eq:avg_quality}
\frac{1}{n}\sum_{H \in \mathcal{H}_{i+1}^{(n)}} q(H) \geq \frac{1}{n}\sum_{H \in \mathcal{H}_{i}^{(n)}} q(H)
\end{equation}
Equation~\eqref{eq:min_quality} ensures the minimum quality is non-decreasing, while Equation~\eqref{eq:avg_quality} guarantees average quality improvement. These properties hold by construction of the TopN operation (Equation~\eqref{eq:topn_hdl}): since $\mathcal{C}_{<i+1} \supset \mathcal{C}_{<i}$, selecting top-$n$ from a strictly larger set that includes all previous top-$n$ elements plus new candidates from layer $i$ cannot decrease either minimum or average quality.

This quality guarantee establishes non-decreasing information quality flow across proposer layers. Consequently, agents in deeper layers operate on strictly superior reference materials, leading to expected quality improvements:
\begin{equation}
\label{eq:expected_quality}
\mathbb{E}[q_{i+1,j}] \geq \mathbb{E}[q_{i,j}], \quad \forall j \in \{1, \ldots, M\}
\end{equation}
where the expectation is taken over LLM generation stochasticity. This monotonicity property prevents information loss during multi-layer process, enabling consistent performance gains as depth increases, in contrast to the degradation observed in standard MoA.

\subsection{Synergistic Enhancement through Multi-Path Generation}

The multi-path generation strategy synergistically addresses both dimensions of MoA performance (Equation~\eqref{eq:moa_performance}): proposer quality and diversity. 
% We analyze how two-stage generation enhances quality through improved agent capability and how heterogeneous path mixing amplifies diversity.

\textbf{Quality Enhancement via Intermediate Representations.} Recall that proposer quality $Q_i$ depends on both agent capability and reference code quality (Equation~\eqref{eq:proposer_quality}). The two-stage mechanism enhances agent capability by leveraging LLMs' substantially stronger parametric knowledge of C++/Python~\cite{xiong2024hlspilot}. In Stage 1, the LLM decomposes abstract design requirements into explicit algorithmic constructs (control flow, data structures, and computational patterns) that directly correspond to hardware behavior~\cite{xu2024automated}. This intermediate representation serves as a structured scaffold for Stage 2 HDL generation, effectively translating the LLM's strong high-level language comprehension into accurate hardware implementations.

To formalize quality propagation in multi-path generation, consider an agent following the $\mathcal{L}$-path at layer $i$. The expected HDL quality depends on both intermediate code quality and reference HDL quality:
\begin{equation}
\label{eq:hdl_quality_decompose}
\mathbb{E}[q(H_{i,j}^{\mathcal{L}})] = \mathbb{E}[g(q(I_{i,j}^{\mathcal{L}}), \text{quality}(\mathcal{H}_{i}^{(n)}))]
\end{equation}
where $g$ represents the Stage 2 translation process, whose expected output quality increases monotonically with both input arguments. Quality-guided caching for intermediate codes ensures:
\begin{equation}
\label{eq:intermediate_quality_bound}
\min_{I \in \mathcal{I}_{i+1}^{\mathcal{L},(k)}} q(I) \geq \min_{I \in \mathcal{I}_{i}^{\mathcal{L},(k)}} q(I)
\end{equation}
\begin{equation}
\label{eq:intermediate_avg_quality}
\frac{1}{k}\sum_{I \in \mathcal{I}_{i+1}^{\mathcal{L},(k)}} q(I) \geq \frac{1}{k}\sum_{I \in \mathcal{I}_{i}^{\mathcal{L},(k)}} q(I)
\end{equation}
Combined with HDL quality guarantees (Equations~\eqref{eq:min_quality}~\eqref{eq:avg_quality}), we have:
\begin{equation}
\label{eq:hdl_quality_bound}
\text{quality}(\mathcal{H}_{i+1}^{(n)}) \geq \text{quality}(\mathcal{H}_{i}^{(n)})
\end{equation}
Since $g$ is expected to be monotonically increasing and both inputs improve, applying the monotonicity yields:
\begin{equation}
\label{eq:multipath_quality_improve}
\begin{aligned}
\mathbb{E}[q(H_{i+1,j}^{\mathcal{L}})] &= \mathbb{E}[g(q(I_{i+1,j}^{\mathcal{L}}), \text{quality}(\mathcal{H}_{i+1}^{(n)}))] \\
&\geq \mathbb{E}[g(q(I_{i,j}^{\mathcal{L}}), \text{quality}(\mathcal{H}_{i}^{(n)}))] = \mathbb{E}[q(H_{i,j}^{\mathcal{L}})]
\end{aligned}
\end{equation}
This monotonicity extends across all agent types (Baseline, C++, Python), collectively ensuring $Q_{i+1} \geq Q_i$.

\textbf{Diversity Amplification through Heterogeneous Paths.} The multi-path architecture amplifies diversity through fundamentally different reasoning trajectories: baseline agents translate directly to HDL, C++ agents reason through imperative abstractions, while Python agents leverage high-level expressiveness. Let $\mathcal{H}_i^{\text{Base}}$, $\mathcal{H}_i^{\text{Cpp}}$, and $\mathcal{H}_i^{\text{Py}}$ denote the HDL outputs from each agent type at layer $i$. The structural diversity is characterized by:
\begin{equation}
\label{eq:path_diversity}
d_i = \mathbb{E}[\text{dissimilarity}(\mathcal{H}_i^{\text{Base}}, \mathcal{H}_i^{\text{Cpp}}, \mathcal{H}_i^{\text{Py}})]
\end{equation}
where dissimilarity captures differences in code structure and implementation patterns, directly enhancing the diversity term $d$ in Equation~\eqref{eq:moa_performance}.

\textbf{Synergistic Performance Maximization.} The heterogeneous mixture of quality-improving paths (Equation~\eqref{eq:multipath_quality_improve}), each exploring distinct solution spaces (Equation~\eqref{eq:path_diversity}), synergistically maximizes MoA performance according to Equation~\eqref{eq:moa_performance}. Quality-guided caching ensures monotonic accumulation of high-quality references across all paths, while multi-path generation expands the solution space through diverse reasoning strategies.

\begin{table*}[!t]
\centering
\footnotesize
\caption{Performance comparison of non-training baselines and our \ourtool on VerilogEval 2.0 and RTLLM 2.0 benchmarks.}
\label{tab:results}
\resizebox{\textwidth}{!}{
\renewcommand{\arraystretch}{0.6}
\setlength{\aboverulesep}{2pt}
\setlength{\belowrulesep}{2pt}
\begin{tabular}{llllllll}
\toprule
\textbf{Model} & \textbf{Method} & \multicolumn{3}{c}{\textbf{VerilogEval 2.0 (\%)}} & \multicolumn{3}{c}{\textbf{RTLLM 2.0 (\%)}} \\
\cmidrule(lr){3-5} \cmidrule(lr){6-8}
& & \textit{Pass@1} & \textit{Pass@3} & \textit{Pass@5} & \textit{Pass@1} & \textit{Pass@3} & \textit{Pass@5} \\
\midrule
\multirow{5}{*}{Qwen2.5-7B} 
& Direct & {\scriptsize 22.90} \phantom{{\tiny \textcolor{green!70!black}{$\uparrow$33.54}}} & {\scriptsize 32.20} \phantom{{\tiny \textcolor{green!70!black}{$\uparrow$29.87}}} & {\scriptsize 40.39} \phantom{{\tiny \textcolor{green!70!black}{$\uparrow$24.88}}} & {\scriptsize 18.99} \phantom{{\tiny \textcolor{green!70!black}{$\uparrow$33.08}}} & {\scriptsize 30.53} \phantom{{\tiny \textcolor{green!70!black}{$\uparrow$27.57}}} & {\scriptsize 36.51} \phantom{{\tiny \textcolor{green!70!black}{$\uparrow$23.95}}} \\
& CoT & {\scriptsize 26.87} {\tiny \textcolor{green!70!black}{$\uparrow$3.97}} & {\scriptsize 35.33} {\tiny \textcolor{green!70!black}{$\uparrow$3.13}} & {\scriptsize 42.58} {\tiny \textcolor{green!70!black}{$\uparrow$2.19}} & {\scriptsize 27.49} {\tiny \textcolor{green!70!black}{$\uparrow$8.50}} & {\scriptsize 34.76} {\tiny \textcolor{green!70!black}{$\uparrow$4.23}} & {\scriptsize 39.95} {\tiny \textcolor{green!70!black}{$\uparrow$3.44}} \\
& HDLCoRe & {\scriptsize 30.92} {\tiny \textcolor{green!70!black}{$\uparrow$8.02}} & {\scriptsize 37.86} {\tiny \textcolor{green!70!black}{$\uparrow$5.66}} & {\scriptsize 46.49} {\tiny \textcolor{green!70!black}{$\uparrow$6.10}} & {\scriptsize 34.68} {\tiny \textcolor{green!70!black}{$\uparrow$15.69}} & {\scriptsize 40.32} {\tiny \textcolor{green!70!black}{$\uparrow$9.79}} & {\scriptsize 44.82} {\tiny \textcolor{green!70!black}{$\uparrow$8.31}} \\
& VeriMaAS & {\scriptsize 32.81} {\tiny \textcolor{green!70!black}{$\uparrow$9.91}} & {\scriptsize 40.25} {\tiny \textcolor{green!70!black}{$\uparrow$8.05}} & {\scriptsize 49.07} {\tiny \textcolor{green!70!black}{$\uparrow$8.68}} & {\scriptsize 40.33} {\tiny \textcolor{green!70!black}{$\uparrow$21.34}} & {\scriptsize 47.24} {\tiny \textcolor{green!70!black}{$\uparrow$16.71}} & {\scriptsize 50.47} {\tiny \textcolor{green!70!black}{$\uparrow$13.96}} \\
& \textbf{\ourtool} & \textbf{\scriptsize 56.44} {\tiny \textcolor{green!70!black}{\textbf{$\uparrow$33.54}}} & \textbf{\scriptsize 62.07} {\tiny \textcolor{green!70!black}{\textbf{$\uparrow$29.87}}} & \textbf{\scriptsize 65.27} {\tiny \textcolor{green!70!black}{\textbf{$\uparrow$24.88}}} & \textbf{\scriptsize 52.07} {\tiny \textcolor{green!70!black}{\textbf{$\uparrow$33.08}}} & \textbf{\scriptsize 58.10} {\tiny \textcolor{green!70!black}{\textbf{$\uparrow$27.57}}} & \textbf{\scriptsize 60.46} {\tiny \textcolor{green!70!black}{\textbf{$\uparrow$23.95}}} \\[1pt]
\hline
\multirow{5}{*}{Qwen2.5-Coder-7B} 
& Direct & {\scriptsize 33.91} \phantom{{\tiny \textcolor{green!70!black}{$\uparrow$27.05}}} & {\scriptsize 43.90} \phantom{{\tiny \textcolor{green!70!black}{$\uparrow$24.23}}} & {\scriptsize 48.06} \phantom{{\tiny \textcolor{green!70!black}{$\uparrow$22.63}}} & {\scriptsize 26.16} \phantom{{\tiny \textcolor{green!70!black}{$\uparrow$28.27}}} & {\scriptsize 36.94} \phantom{{\tiny \textcolor{green!70!black}{$\uparrow$25.58}}} & {\scriptsize 41.11} \phantom{{\tiny \textcolor{green!70!black}{$\uparrow$25.13}}} \\
& CoT & {\scriptsize 36.74} {\tiny \textcolor{green!70!black}{$\uparrow$2.83}} & {\scriptsize 45.88} {\tiny \textcolor{green!70!black}{$\uparrow$1.98}} & {\scriptsize 50.99} {\tiny \textcolor{green!70!black}{$\uparrow$2.93}} & {\scriptsize 31.87} {\tiny \textcolor{green!70!black}{$\uparrow$5.71}} & {\scriptsize 39.36} {\tiny \textcolor{green!70!black}{$\uparrow$2.42}} & {\scriptsize 45.36} {\tiny \textcolor{green!70!black}{$\uparrow$4.25}} \\
& HDLCoRe & {\scriptsize 40.67} {\tiny \textcolor{green!70!black}{$\uparrow$6.76}} & {\scriptsize 47.81} {\tiny \textcolor{green!70!black}{$\uparrow$3.91}} & {\scriptsize 52.64} {\tiny \textcolor{green!70!black}{$\uparrow$4.58}} & {\scriptsize 36.26} {\tiny \textcolor{green!70!black}{$\uparrow$10.10}} & {\scriptsize 45.61} {\tiny \textcolor{green!70!black}{$\uparrow$8.67}} & {\scriptsize 51.12} {\tiny \textcolor{green!70!black}{$\uparrow$10.01}} \\
& VeriMaAS & {\scriptsize 44.73} {\tiny \textcolor{green!70!black}{$\uparrow$10.82}} & {\scriptsize 51.62} {\tiny \textcolor{green!70!black}{$\uparrow$7.72}} & {\scriptsize 55.78} {\tiny \textcolor{green!70!black}{$\uparrow$7.72}} & {\scriptsize 44.71} {\tiny \textcolor{green!70!black}{$\uparrow$18.55}} & {\scriptsize 51.37} {\tiny \textcolor{green!70!black}{$\uparrow$14.43}} & {\scriptsize 56.46} {\tiny \textcolor{green!70!black}{$\uparrow$15.35}} \\
& \textbf{\ourtool} & \textbf{\scriptsize 60.96} {\tiny \textcolor{green!70!black}{\textbf{$\uparrow$27.05}}} & \textbf{\scriptsize 68.13} {\tiny \textcolor{green!70!black}{\textbf{$\uparrow$24.23}}} & \textbf{\scriptsize 70.69} {\tiny \textcolor{green!70!black}{\textbf{$\uparrow$22.63}}} & \textbf{\scriptsize 54.43} {\tiny \textcolor{green!70!black}{\textbf{$\uparrow$28.27}}} & \textbf{\scriptsize 62.52} {\tiny \textcolor{green!70!black}{\textbf{$\uparrow$25.58}}} & \textbf{\scriptsize 66.24} {\tiny \textcolor{green!70!black}{\textbf{$\uparrow$25.13}}} \\[1pt]
\hline
\multirow{5}{*}{Qwen2.5-14B} 
& Direct & {\scriptsize 39.26} \phantom{{\tiny \textcolor{green!70!black}{$\uparrow$26.01}}} & {\scriptsize 49.98} \phantom{{\tiny \textcolor{green!70!black}{$\uparrow$21.63}}} & {\scriptsize 53.74} \phantom{{\tiny \textcolor{green!70!black}{$\uparrow$20.90}}} & {\scriptsize 31.71} \phantom{{\tiny \textcolor{green!70!black}{$\uparrow$23.35}}} & {\scriptsize 40.20} \phantom{{\tiny \textcolor{green!70!black}{$\uparrow$22.56}}} & {\scriptsize 45.87} \phantom{{\tiny \textcolor{green!70!black}{$\uparrow$20.46}}} \\
& CoT & {\scriptsize 41.24} {\tiny \textcolor{green!70!black}{$\uparrow$1.98}} & {\scriptsize 51.62} {\tiny \textcolor{green!70!black}{$\uparrow$1.64}} & {\scriptsize 54.37} {\tiny \textcolor{green!70!black}{$\uparrow$0.63}} & {\scriptsize 36.49} {\tiny \textcolor{green!70!black}{$\uparrow$4.78}} & {\scriptsize 44.64} {\tiny \textcolor{green!70!black}{$\uparrow$4.44}} & {\scriptsize 48.29} {\tiny \textcolor{green!70!black}{$\uparrow$2.42}} \\
& HDLCoRe & {\scriptsize 45.18} {\tiny \textcolor{green!70!black}{$\uparrow$5.92}} & {\scriptsize 55.83} {\tiny \textcolor{green!70!black}{$\uparrow$5.85}} & {\scriptsize 59.42} {\tiny \textcolor{green!70!black}{$\uparrow$5.68}} & {\scriptsize 41.78} {\tiny \textcolor{green!70!black}{$\uparrow$10.07}} & {\scriptsize 49.97} {\tiny \textcolor{green!70!black}{$\uparrow$9.77}} & {\scriptsize 53.17} {\tiny \textcolor{green!70!black}{$\uparrow$7.30}} \\
& VeriMaAS & {\scriptsize 48.97} {\tiny \textcolor{green!70!black}{$\uparrow$9.71}} & {\scriptsize 57.84} {\tiny \textcolor{green!70!black}{$\uparrow$7.86}} & {\scriptsize 61.77} {\tiny \textcolor{green!70!black}{$\uparrow$8.03}} & {\scriptsize 48.14} {\tiny \textcolor{green!70!black}{$\uparrow$16.43}} & {\scriptsize 54.28} {\tiny \textcolor{green!70!black}{$\uparrow$14.08}} & {\scriptsize 58.76} {\tiny \textcolor{green!70!black}{$\uparrow$12.89}} \\
& \textbf{\ourtool} & \textbf{\scriptsize 65.27} {\tiny \textcolor{green!70!black}{\textbf{$\uparrow$26.01}}} & \textbf{\scriptsize 71.61} {\tiny \textcolor{green!70!black}{\textbf{$\uparrow$21.63}}} & \textbf{\scriptsize 74.64} {\tiny \textcolor{green!70!black}{\textbf{$\uparrow$20.90}}} & \textbf{\scriptsize 55.06} {\tiny \textcolor{green!70!black}{\textbf{$\uparrow$23.35}}} & \textbf{\scriptsize 62.76} {\tiny \textcolor{green!70!black}{\textbf{$\uparrow$22.56}}} & \textbf{\scriptsize 66.33} {\tiny \textcolor{green!70!black}{\textbf{$\uparrow$20.46}}} \\[1pt]
\hline
\multirow{5}{*}{Qwen2.5-Coder-14B} 
& Direct & {\scriptsize 39.74} \phantom{{\tiny \textcolor{green!70!black}{$\uparrow$27.12}}} & {\scriptsize 49.72} \phantom{{\tiny \textcolor{green!70!black}{$\uparrow$23.52}}} & {\scriptsize 52.21} \phantom{{\tiny \textcolor{green!70!black}{$\uparrow$24.17}}} & {\scriptsize 35.61} \phantom{{\tiny \textcolor{green!70!black}{$\uparrow$24.15}}} & {\scriptsize 42.34} \phantom{{\tiny \textcolor{green!70!black}{$\uparrow$21.99}}} & {\scriptsize 46.43} \phantom{{\tiny \textcolor{green!70!black}{$\uparrow$20.79}}} \\
& CoT & {\scriptsize 43.81} {\tiny \textcolor{green!70!black}{$\uparrow$4.07}} & {\scriptsize 51.27} {\tiny \textcolor{green!70!black}{$\uparrow$1.55}} & {\scriptsize 54.11} {\tiny \textcolor{green!70!black}{$\uparrow$1.90}} & {\scriptsize 40.65} {\tiny \textcolor{green!70!black}{$\uparrow$5.04}} & {\scriptsize 48.77} {\tiny \textcolor{green!70!black}{$\uparrow$6.43}} & {\scriptsize 51.72} {\tiny \textcolor{green!70!black}{$\uparrow$5.29}} \\
& HDLCoRe & {\scriptsize 46.82} {\tiny \textcolor{green!70!black}{$\uparrow$7.08}} & {\scriptsize 54.79} {\tiny \textcolor{green!70!black}{$\uparrow$5.07}} & {\scriptsize 57.04} {\tiny \textcolor{green!70!black}{$\uparrow$4.83}} & {\scriptsize 47.86} {\tiny \textcolor{green!70!black}{$\uparrow$12.25}} & {\scriptsize 55.19} {\tiny \textcolor{green!70!black}{$\uparrow$12.85}} & {\scriptsize 58.18} {\tiny \textcolor{green!70!black}{$\uparrow$11.75}} \\
& VeriMaAS & {\scriptsize 50.96} {\tiny \textcolor{green!70!black}{$\uparrow$11.22}} & {\scriptsize 58.48} {\tiny \textcolor{green!70!black}{$\uparrow$8.76}} & {\scriptsize 62.63} {\tiny \textcolor{green!70!black}{$\uparrow$10.42}} & {\scriptsize 51.20} {\tiny \textcolor{green!70!black}{$\uparrow$15.59}} & {\scriptsize 58.93} {\tiny \textcolor{green!70!black}{$\uparrow$16.59}} & {\scriptsize 61.67} {\tiny \textcolor{green!70!black}{$\uparrow$15.24}} \\
& \textbf{\ourtool} & \textbf{\scriptsize 66.86} {\tiny \textcolor{green!70!black}{\textbf{$\uparrow$27.12}}} & \textbf{\scriptsize 73.24} {\tiny \textcolor{green!70!black}{\textbf{$\uparrow$23.52}}} & \textbf{\scriptsize 76.38} {\tiny \textcolor{green!70!black}{\textbf{$\uparrow$24.17}}} & \textbf{\scriptsize 59.76} {\tiny \textcolor{green!70!black}{\textbf{$\uparrow$24.15}}} & \textbf{\scriptsize 64.33} {\tiny \textcolor{green!70!black}{\textbf{$\uparrow$21.99}}} & \textbf{\scriptsize 67.22} {\tiny \textcolor{green!70!black}{\textbf{$\uparrow$20.79}}} \\[1pt]
\hline
\multirow{5}{*}{Qwen2.5-32B} 
& Direct & {\scriptsize 46.85} \phantom{{\tiny \textcolor{green!70!black}{$\uparrow$25.00}}} & {\scriptsize 56.11} \phantom{{\tiny \textcolor{green!70!black}{$\uparrow$20.37}}} & {\scriptsize 58.41} \phantom{{\tiny \textcolor{green!70!black}{$\uparrow$19.75}}} & {\scriptsize 43.62} \phantom{{\tiny \textcolor{green!70!black}{$\uparrow$20.19}}} & {\scriptsize 48.73} \phantom{{\tiny \textcolor{green!70!black}{$\uparrow$18.87}}} & {\scriptsize 50.75} \phantom{{\tiny \textcolor{green!70!black}{$\uparrow$19.13}}} \\
& CoT & {\scriptsize 48.72} {\tiny \textcolor{green!70!black}{$\uparrow$1.87}} & {\scriptsize 57.25} {\tiny \textcolor{green!70!black}{$\uparrow$1.14}} & {\scriptsize 59.80} {\tiny \textcolor{green!70!black}{$\uparrow$1.39}} & {\scriptsize 46.99} {\tiny \textcolor{green!70!black}{$\uparrow$3.37}} & {\scriptsize 50.50} {\tiny \textcolor{green!70!black}{$\uparrow$1.77}} & {\scriptsize 52.46} {\tiny \textcolor{green!70!black}{$\uparrow$1.71}} \\
& HDLCoRe & {\scriptsize 51.75} {\tiny \textcolor{green!70!black}{$\uparrow$4.90}} & {\scriptsize 59.63} {\tiny \textcolor{green!70!black}{$\uparrow$3.52}} & {\scriptsize 61.79} {\tiny \textcolor{green!70!black}{$\uparrow$3.38}} & {\scriptsize 50.95} {\tiny \textcolor{green!70!black}{$\uparrow$7.33}} & {\scriptsize 54.06} {\tiny \textcolor{green!70!black}{$\uparrow$5.33}} & {\scriptsize 58.98} {\tiny \textcolor{green!70!black}{$\uparrow$8.23}} \\
& VeriMaAS & {\scriptsize 53.57} {\tiny \textcolor{green!70!black}{$\uparrow$6.72}} & {\scriptsize 61.82} {\tiny \textcolor{green!70!black}{$\uparrow$5.71}} & {\scriptsize 64.26} {\tiny \textcolor{green!70!black}{$\uparrow$5.85}} & {\scriptsize 54.18} {\tiny \textcolor{green!70!black}{$\uparrow$10.56}} & {\scriptsize 57.21} {\tiny \textcolor{green!70!black}{$\uparrow$8.48}} & {\scriptsize 61.54} {\tiny \textcolor{green!70!black}{$\uparrow$10.79}} \\
& \textbf{\ourtool} & \textbf{\scriptsize 71.85} {\tiny \textcolor{green!70!black}{\textbf{$\uparrow$25.00}}} & \textbf{\scriptsize 76.48} {\tiny \textcolor{green!70!black}{\textbf{$\uparrow$20.37}}} & \textbf{\scriptsize 78.16} {\tiny \textcolor{green!70!black}{\textbf{$\uparrow$19.75}}} & \textbf{\scriptsize 63.81} {\tiny \textcolor{green!70!black}{\textbf{$\uparrow$20.19}}} & \textbf{\scriptsize 67.60} {\tiny \textcolor{green!70!black}{\textbf{$\uparrow$18.87}}} & \textbf{\scriptsize 69.88} {\tiny \textcolor{green!70!black}{\textbf{$\uparrow$19.13}}} \\[1pt]
\hline
\multirow{5}{*}{Qwen2.5-Coder-32B} 
& Direct & {\scriptsize 46.93} \phantom{{\tiny \textcolor{green!70!black}{$\uparrow$26.38}}} & {\scriptsize 55.73} \phantom{{\tiny \textcolor{green!70!black}{$\uparrow$23.32}}} & {\scriptsize 57.12} \phantom{{\tiny \textcolor{green!70!black}{$\uparrow$24.15}}} & {\scriptsize 40.40} \phantom{{\tiny \textcolor{green!70!black}{$\uparrow$25.09}}} & {\scriptsize 45.42} \phantom{{\tiny \textcolor{green!70!black}{$\uparrow$23.45}}} & {\scriptsize 48.24} \phantom{{\tiny \textcolor{green!70!black}{$\uparrow$22.00}}} \\
& CoT & {\scriptsize 48.65} {\tiny \textcolor{green!70!black}{$\uparrow$1.72}} & {\scriptsize 56.31} {\tiny \textcolor{green!70!black}{$\uparrow$0.58}} & {\scriptsize 59.34} {\tiny \textcolor{green!70!black}{$\uparrow$2.22}} & {\scriptsize 45.72} {\tiny \textcolor{green!70!black}{$\uparrow$5.32}} & {\scriptsize 48.75} {\tiny \textcolor{green!70!black}{$\uparrow$3.33}} & {\scriptsize 50.29} {\tiny \textcolor{green!70!black}{$\uparrow$2.05}} \\
& HDLCoRe & {\scriptsize 51.28} {\tiny \textcolor{green!70!black}{$\uparrow$4.35}} & {\scriptsize 58.63} {\tiny \textcolor{green!70!black}{$\uparrow$2.90}} & {\scriptsize 61.47} {\tiny \textcolor{green!70!black}{$\uparrow$4.35}} & {\scriptsize 51.72} {\tiny \textcolor{green!70!black}{$\uparrow$11.32}} & {\scriptsize 54.73} {\tiny \textcolor{green!70!black}{$\uparrow$9.31}} & {\scriptsize 59.64} {\tiny \textcolor{green!70!black}{$\uparrow$11.40}} \\
& VeriMaAS & {\scriptsize 56.67} {\tiny \textcolor{green!70!black}{$\uparrow$9.74}} & {\scriptsize 63.46} {\tiny \textcolor{green!70!black}{$\uparrow$7.73}} & {\scriptsize 66.92} {\tiny \textcolor{green!70!black}{$\uparrow$9.80}} & {\scriptsize 55.82} {\tiny \textcolor{green!70!black}{$\uparrow$15.42}} & {\scriptsize 59.97} {\tiny \textcolor{green!70!black}{$\uparrow$14.55}} & {\scriptsize 62.31} {\tiny \textcolor{green!70!black}{$\uparrow$14.07}} \\
& \textbf{\ourtool} & \textbf{\scriptsize 73.31} {\tiny \textcolor{green!70!black}{\textbf{$\uparrow$26.38}}} & \textbf{\scriptsize 79.05} {\tiny \textcolor{green!70!black}{\textbf{$\uparrow$23.32}}} & \textbf{\scriptsize 81.27} {\tiny \textcolor{green!70!black}{\textbf{$\uparrow$24.15}}} & \textbf{\scriptsize 65.49} {\tiny \textcolor{green!70!black}{\textbf{$\uparrow$25.09}}} & \textbf{\scriptsize 71.42} {\tiny \textcolor{green!70!black}{\textbf{$\uparrow$26.00}}} & \textbf{\scriptsize 74.11} {\tiny \textcolor{green!70!black}{\textbf{$\uparrow$25.87}}} \\[1pt]
\hline
\multirow{5}{*}{GPT-4o-mini} 
& Direct & {\scriptsize 48.97} \phantom{{\tiny \textcolor{green!70!black}{$\uparrow$23.46}}} & {\scriptsize 56.94} \phantom{{\tiny \textcolor{green!70!black}{$\uparrow$20.00}}} & {\scriptsize 58.62} \phantom{{\tiny \textcolor{green!70!black}{$\uparrow$20.84}}} & {\scriptsize 46.60} \phantom{{\tiny \textcolor{green!70!black}{$\uparrow$17.63}}} & {\scriptsize 50.71} \phantom{{\tiny \textcolor{green!70!black}{$\uparrow$16.74}}} & {\scriptsize 51.95} \phantom{{\tiny \textcolor{green!70!black}{$\uparrow$16.72}}} \\
& CoT & {\scriptsize 52.20} {\tiny \textcolor{green!70!black}{$\uparrow$3.23}} & {\scriptsize 59.83} {\tiny \textcolor{green!70!black}{$\uparrow$2.89}} & {\scriptsize 60.93} {\tiny \textcolor{green!70!black}{$\uparrow$2.31}} & {\scriptsize 48.27} {\tiny \textcolor{green!70!black}{$\uparrow$1.67}} & {\scriptsize 53.56} {\tiny \textcolor{green!70!black}{$\uparrow$2.85}} & {\scriptsize 55.48} {\tiny \textcolor{green!70!black}{$\uparrow$3.53}} \\
& HDLCoRe & {\scriptsize 53.50} {\tiny \textcolor{green!70!black}{$\uparrow$4.53}} & {\scriptsize 61.46} {\tiny \textcolor{green!70!black}{$\uparrow$4.52}} & {\scriptsize 63.72} {\tiny \textcolor{green!70!black}{$\uparrow$5.10}} & {\scriptsize 51.19} {\tiny \textcolor{green!70!black}{$\uparrow$4.59}} & {\scriptsize 56.72} {\tiny \textcolor{green!70!black}{$\uparrow$6.01}} & {\scriptsize 59.01} {\tiny \textcolor{green!70!black}{$\uparrow$7.06}} \\
& VeriMaAS & {\scriptsize 57.24} {\tiny \textcolor{green!70!black}{$\uparrow$8.27}} & {\scriptsize 64.77} {\tiny \textcolor{green!70!black}{$\uparrow$7.83}} & {\scriptsize 66.85} {\tiny \textcolor{green!70!black}{$\uparrow$8.23}} & {\scriptsize 57.25} {\tiny \textcolor{green!70!black}{$\uparrow$10.65}} & {\scriptsize 61.46} {\tiny \textcolor{green!70!black}{$\uparrow$10.75}} & {\scriptsize 63.17} {\tiny \textcolor{green!70!black}{$\uparrow$11.22}} \\
& \textbf{\ourtool} & \textbf{\scriptsize 72.43} {\tiny \textcolor{green!70!black}{\textbf{$\uparrow$23.46}}} & \textbf{\scriptsize 76.94} {\tiny \textcolor{green!70!black}{\textbf{$\uparrow$20.00}}} & \textbf{\scriptsize 79.46} {\tiny \textcolor{green!70!black}{\textbf{$\uparrow$20.84}}} & \textbf{\scriptsize 64.23} {\tiny \textcolor{green!70!black}{\textbf{$\uparrow$17.63}}} & \textbf{\scriptsize 67.45} {\tiny \textcolor{green!70!black}{\textbf{$\uparrow$16.74}}} & \textbf{\scriptsize 68.67} {\tiny \textcolor{green!70!black}{\textbf{$\uparrow$16.72}}} \\[1pt]
\hline
\multirow{5}{*}{GPT-4o} 
& Direct & {\scriptsize 64.74} \phantom{{\tiny \textcolor{green!70!black}{$\uparrow$20.23}}} & {\scriptsize 69.89} \phantom{{\tiny \textcolor{green!70!black}{$\uparrow$19.76}}} & {\scriptsize 71.66} \phantom{{\tiny \textcolor{green!70!black}{$\uparrow$19.37}}} & {\scriptsize 52.48} \phantom{{\tiny \textcolor{green!70!black}{$\uparrow$16.69}}} & {\scriptsize 56.18} \phantom{{\tiny \textcolor{green!70!black}{$\uparrow$17.50}}} & {\scriptsize 57.62} \phantom{{\tiny \textcolor{green!70!black}{$\uparrow$18.00}}} \\
& CoT & {\scriptsize 66.38} {\tiny \textcolor{green!70!black}{$\uparrow$1.64}} & {\scriptsize 71.18} {\tiny \textcolor{green!70!black}{$\uparrow$1.29}} & {\scriptsize 74.22} {\tiny \textcolor{green!70!black}{$\uparrow$2.56}} & {\scriptsize 54.89} {\tiny \textcolor{green!70!black}{$\uparrow$2.41}} & {\scriptsize 59.22} {\tiny \textcolor{green!70!black}{$\uparrow$3.04}} & {\scriptsize 61.77} {\tiny \textcolor{green!70!black}{$\uparrow$4.15}} \\
& HDLCoRe & {\scriptsize 69.60} {\tiny \textcolor{green!70!black}{$\uparrow$4.86}} & {\scriptsize 73.53} {\tiny \textcolor{green!70!black}{$\uparrow$3.64}} & {\scriptsize 75.86} {\tiny \textcolor{green!70!black}{$\uparrow$4.20}} & {\scriptsize 56.27} {\tiny \textcolor{green!70!black}{$\uparrow$3.79}} & {\scriptsize 61.47} {\tiny \textcolor{green!70!black}{$\uparrow$5.29}} & {\scriptsize 63.23} {\tiny \textcolor{green!70!black}{$\uparrow$5.61}} \\
& VeriMaAS & {\scriptsize 71.34} {\tiny \textcolor{green!70!black}{$\uparrow$6.60}} & {\scriptsize 76.12} {\tiny \textcolor{green!70!black}{$\uparrow$6.23}} & {\scriptsize 79.21} {\tiny \textcolor{green!70!black}{$\uparrow$7.55}} & {\scriptsize 61.70} {\tiny \textcolor{green!70!black}{$\uparrow$9.22}} & {\scriptsize 67.25} {\tiny \textcolor{green!70!black}{$\uparrow$11.07}} & {\scriptsize 68.84} {\tiny \textcolor{green!70!black}{$\uparrow$11.22}} \\
& \textbf{\ourtool} & \textbf{\scriptsize 84.97} {\tiny \textcolor{green!70!black}{\textbf{$\uparrow$20.23}}} & \textbf{\scriptsize 89.65} {\tiny \textcolor{green!70!black}{\textbf{$\uparrow$19.76}}} & \textbf{\scriptsize 91.03} {\tiny \textcolor{green!70!black}{\textbf{$\uparrow$19.37}}} & \textbf{\scriptsize 69.17} {\tiny \textcolor{green!70!black}{\textbf{$\uparrow$16.69}}} & \textbf{\scriptsize 73.68} {\tiny \textcolor{green!70!black}{\textbf{$\uparrow$17.50}}} & \textbf{\scriptsize 75.62} {\tiny \textcolor{green!70!black}{\textbf{$\uparrow$18.00}}} \\
\bottomrule
\end{tabular}
}
\end{table*}

\section{Experiments}

We conduct comprehensive experiments to evaluate the effectiveness of our \ourtool framework. Our evaluation is designed to address the following key research questions:

\textbf{RQ1:} How does \ourtool compare to state-of-the-art methods in HDL generation performance?

\textbf{RQ2:} What is the contribution of each core component to the overall performance?

\textbf{RQ3:} How sensitive is the framework to variations in key parameters?

\textbf{RQ4:} What is the inference cost, and how does \ourtool compare to baselines under ISO-cost conditions?

\textbf{RQ5:} How robust is \ourtool when golden testbenches are unavailable, and what are the failure modes of the two-stage process?

\textbf{RQ6:} Can \ourtool effectively explore diverse solutions and propagate high-quality information across layers?

\subsection{Experimental Setup}

\textbf{Benchmarks.} We evaluate on VerilogEval 2.0~\cite{pinckney2025revisiting} (156 problems covering combinational and sequential circuits) and RTLLM 2.0~\cite{liu2024openllm} (50 complex real-world design tasks).

\textbf{Baselines.} We compare \ourtool against representative methods from two categories. For non-training approaches, we include Direct (direct generation without prompting strategies), CoT~\cite{wei2022chain} (Chain-of-Thought prompting), HDLCoRe~\cite{ping2025hdlcore} (difficulty-adaptive prompting with RAG), and VeriMaAS~\cite{bhattaram2025verimaas} (adaptive multi-agent system), evaluated across various LLMs including Qwen2.5 series and GPT-4o models. For fine-tuned models, we include RTLCoder-Mistral~\cite{liu2024rtlcoder}, RTLCoder-DeepSeek-Coder~\cite{liu2024rtlcoder}, OriGen-DeepSeek-Coder-7B-v1.5~\cite{cui2024origen}, HaVen-CodeQwen1.5~\cite{yang2025haven}, VeriRL-DeepSeek-Coder~\cite{teng2025verirl}, and VeriRL-CodeQwen2.5~\cite{teng2025verirl}.

\begin{table*}[!t]
\centering
\footnotesize
\caption{Performance comparison of fine-tuned Verilog-specific models and our \ourtool on VerilogEval 2.0 and RTLLM 2.0 benchmarks. \colorbox{yellow!50}{First}, \colorbox{cyan!30}{second}, and \colorbox{orange!40}{third} best results are highlighted for each metric.}
\label{tab:finetune}
\resizebox{\textwidth}{!}{
\renewcommand{\arraystretch}{0.85}
\setlength{\aboverulesep}{2pt}
\setlength{\belowrulesep}{2pt}
\begin{tabular}{llccccccc}
\toprule
\multicolumn{2}{c}{\textbf{Model}} & \textbf{Size} & \multicolumn{3}{c}{\textbf{VerilogEval 2.0 (\%)}} & \multicolumn{3}{c}{\textbf{RTLLM 2.0 (\%)}} \\
\cmidrule(lr){4-6} \cmidrule(lr){7-9}
& & & \textit{Pass@1} & \textit{Pass@3} & \textit{Pass@5} & \textit{Pass@1} & \textit{Pass@3} & \textit{Pass@5} \\
\midrule
\multirow{6}{*}{\shortstack[l]{Verilog-Specific\\(finetune)}} 
& RTLCoder-Mistral & 7B & 35.62 & 36.63 & 37.71 & 38.68 & 40.86 & 41.84 \\
& RTLCoder-DeepSeek-Coder & 6.7B & 39.67 & 43.66 & 45.99 & 40.75 & 46.94 & 49.83 \\
& OriGen-DeepSeek-Coder-7B-v1.5 & 7B & 51.19 & 54.88 & 56.78 & 41.11 & 52.09 & 58.48 \\
& HaVen-CodeQwen1.5 & 7B & 55.40 & 59.73 & 62.76 & 51.04 & 57.38 & 60.89 \\
& VeriRL-DeepSeek-Coder & 6.7B & 64.57 & 68.96 & 71.78 & 58.64 & 63.92 & 66.26 \\
& VeriRL-CodeQwen2.5 & 7B & \cellcolor{orange!40}66.28 & \cellcolor{orange!40}71.35 & \cellcolor{orange!40}73.43 & \cellcolor{orange!40}61.53 & \cellcolor{orange!40}65.27 & \cellcolor{orange!40}68.94 \\
\hline
\multirow{4}{*}{Ours} 
& Qwen2.5-Coder-7B + \ourtool & 7B & 60.96 & 68.13 & 70.69 & 54.43 & 62.52 & 66.24 \\
& Qwen2.5-Coder-14B + \ourtool & 14B & 66.86 & 73.24 & 76.38 & 59.76 & 64.33 & 66.22 \\
& Qwen2.5-Coder-32B + \ourtool & 32B & \cellcolor{cyan!30}73.31 & \cellcolor{cyan!30}79.05 & \cellcolor{cyan!30}81.27 & \cellcolor{cyan!30}65.49 & \cellcolor{cyan!30}71.42 & \cellcolor{cyan!30}74.11 \\
& VeriRL-CodeQwen2.5 + \ourtool & 7B & \cellcolor{yellow!50}82.47 & \cellcolor{yellow!50}87.91 & \cellcolor{yellow!50}90.68 & \cellcolor{yellow!50}74.45 & \cellcolor{yellow!50}78.63 & \cellcolor{yellow!50}82.02 \\
\bottomrule
\end{tabular}
}
\end{table*}

\begin{figure*}[t]
  \centering
  \includegraphics[width=\textwidth]{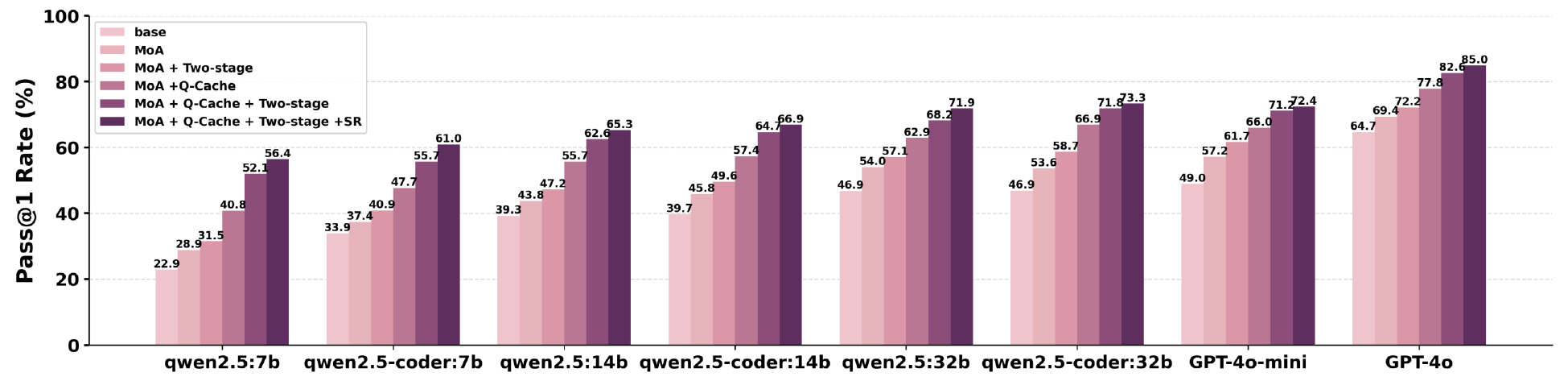}
  \caption{Ablation Study of techniques adopted in our framework. Pass@1 performance [Higher is Better] reported on VerilogEval 2.0 dataset.}
  \label{fig:ablation}
\end{figure*}

\textbf{Evaluation Metric.} We evaluate HDL generation performance from the perspective of functional correctness using the widely adopted pass@k metric~\cite{pinckney2025revisiting}, which estimates the proportion of problems for which at least one of $k$ generated solutions passes functional verification. The metric is defined as:
\begin{equation}
\label{eq:passk}
\text{pass@}k := \mathbb{E}\left[1 - \frac{\binom{n-c}{k}}{\binom{n}{k}}\right]
\end{equation}
where $n \geq k$ is the number of sampled outputs per problem and $c$ is the number of correct outputs among them. Following the experimental setting in VerilogEval 2.0~\cite{pinckney2025revisiting}, we set $n=10$ in all our experiments.

\textbf{Implementation Details.} We use Icarus Verilog (iverilog) as our simulator for its efficiency and ease of integration. For LLM generation, we adopt the sampling strategy with temperature=0.8 and top\_p=0.95, following the configuration in VerilogEval 2.0~\cite{pinckney2025revisiting} to ensure fair comparison. Our \ourtool framework uses 4 proposer layers with 6 parallel agents per layer, consisting of 2 baseline agents, 2 C++ agents, and 2 Python agents, consistent with the standard MoA configuration~\cite{wang2024mixture}. The code is available at: \url{https://github.com/hping666/HDL_Generation}.

\subsection{HDL Generation Performance (RQ1)}

We evaluate \ourtool's HDL generation performance against state-of-the-art baselines on VerilogEval 2.0 and RTLLM 2.0 benchmarks. Tables~\ref{tab:results} and~\ref{tab:finetune} present comprehensive results across multiple LLM backbones and baseline methods.

\textbf{Comparison with Non-Training Methods.} Table~\ref{tab:results} demonstrates that \ourtool consistently outperforms all non-training baselines across different LLM backbones and both benchmarks, with particularly significant improvements on smaller LLMs. On VerilogEval 2.0, \ourtool with Qwen2.5-7B achieves 56.44\% Pass@1, outperforming Direct (22.90\%) by 33.54 points and VeriMaAS (32.81\%) by 23.63 points. Remarkably, \ourtool with Qwen2.5-7B (56.44\%) surpasses VeriMaAS with the larger Qwen2.5-32B (53.57\%), and \ourtool with Qwen2.5-Coder-7B (60.96\%) exceeds VeriMaAS with Qwen2.5-Coder-32B (56.67\%), showing that our framework enables smaller LLMs to match and even exceed larger models through superior information propagation and solution space exploration. At larger scales with Qwen2.5-Coder-32B, \ourtool achieves 73.31\% versus VeriMaAS's 56.67\% (16.64 points improvement). Performance gains extend to commercial models: GPT-4o-mini reaches 72.43\% versus VeriMaAS's 56.24\% (16.19 points improvement), and GPT-4o attains 84.97\% versus 71.34\% (13.63 points improvement). On RTLLM 2.0, \ourtool consistently outperforms VeriMaAS by 7--12 points of Pass@1 across model scales, with stronger gains for smaller models. These results validate that our quality-guided caching and multi-path generation strategies enchance HDL generation performance through architectural innovations applicable to diverse LLMs.

To further validate that \ourtool's improvements are not simply attributable to iterative optimization, we compare against a Simple Hill-Climbing with Mutation baseline using an identical computational budget (4 iterations with 6 candidates per iteration, GPT-4o-mini backbone). Hill-Climbing achieves 58.14\% Pass@1 on VerilogEval 2.0 and 55.47\% on RTLLM 2.0, while \ourtool achieves 72.43\% and 64.23\% respectively---representing gaps of 14.29\% and 8.76\%. This demonstrates that \ourtool's gains stem from its architectural design (reliable quality ranking, global caching to avoid local optima, and multi-path diversity) rather than simple iterative search.

\textbf{Comparison with Fine-Tuned Models.} Table~\ref{tab:finetune} shows that \ourtool achieves competitive or superior performance without requiring training. With Qwen2.5-Coder-7B, \ourtool attains 60.96\% Pass@1 on VerilogEval 2.0, approaching fine-tuned models like VeriRL-DeepSeek-Coder (64.57\%) and demonstrating that effective architectural design can partially compensate for domain-specific training. With Qwen2.5-Coder-14B, \ourtool achieves 66.86\%, matching the best fine-tuned 7B model VeriRL-CodeQwen2.5 (66.28\%). Using Qwen2.5-Coder-32B, \ourtool reaches 73.31\%, surpassing all fine-tuned models by substantial margins (7.03 points improvement over VeriRL-CodeQwen2.5). This demonstrates our training-free framework can match or exceed fine-tuned models through quality-guided progress and diverse solution exploration.

Furthermore, we evaluate whether \ourtool provides complementary benefits when applied on top of fine-tuned models. Using VeriRL-CodeQwen2.5 as the backbone model within \ourtool, we observe improvements of +16.19\% Pass@1 on VerilogEval 2.0 (66.28\% $\to$ 82.47\%) and +12.92\% on RTLLM 2.0 (61.53\% $\to$ 74.45\%), as shown in Table~\ref{tab:finetune}. These results demonstrate that \ourtool and fine-tuned models are complementary rather than redundant, suggesting that \ourtool can continue to boost performance as stronger RTL-specialized models emerge.

\begin{figure}[t]
  \centering
  \includegraphics[width=\columnwidth]{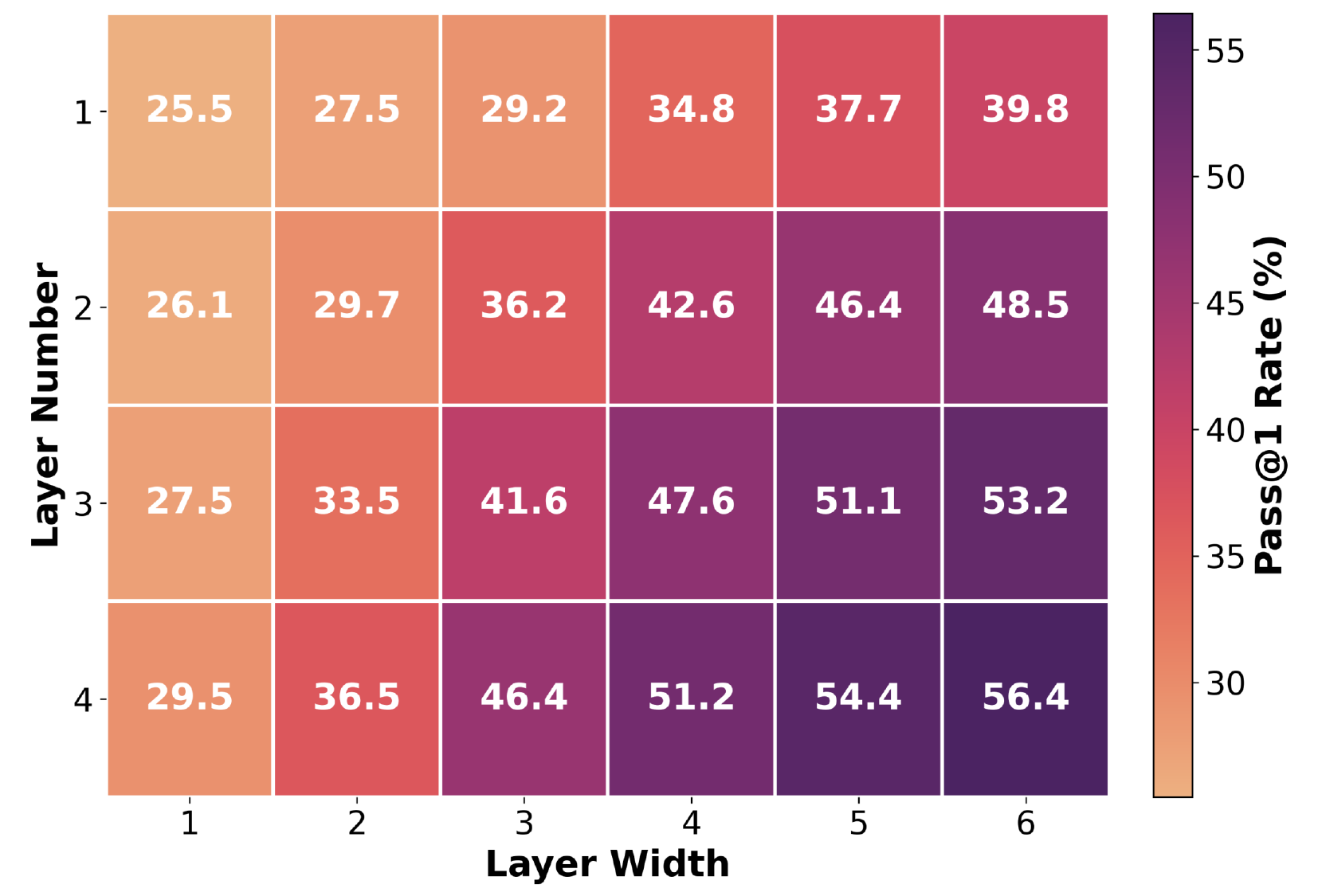}
  \caption{VeriMoA performance (Pass@1)[Darker is Better] on VerilogEval 2.0 using Qwen2.5:7b with varying layer configurations.}
  \label{fig:heatmap}
\end{figure}

\subsection{Ablation Study (RQ2)}
\label{sec:ablation}

We conduct ablation studies to analyze the contribution of each core component. We evaluate six configurations across multiple LLM backbones using Pass@1 on VerilogEval 2.0 basesd on the following techniques: \textbf{Base} (direct generation), \textbf{MoA} (standard Mixture-of-Agents), \textbf{Two-stage} (high-level code as intermediate representations), \textbf{Q-Cache} (quality-guided caching mechanism), and \textbf{SR} (simulator-based self-refinement). Figure~\ref{fig:ablation} reveals that Q-Cache provides substantially greater benefits than Two-stage alone. On Qwen2.5-7B, adding Q-Cache to MoA improves Pass@1 from 28.86\% to 40.79\% (11.93 points improvement), while adding Two-stage to MoA only yields 31.46\% (2.60 points improvement). This pattern holds across all models: with GPT-4o, Q-Cache contributes 8.43 points improvement versus Two-stage's 2.78 points improvement.

More importantly, Q-Cache is essential for realizing Two-stage's full potential. MoA + Two-stage achieves modest gains, but MoA + Q-Cache + Two-stage shows substantial improvements. For Qwen2.5-7B, adding Two-stage on top of Q-Cache improves from 40.79\% to 52.06\% (11.27 points), while adding Two-stage to baseline MoA yields only 2.60 points improvement, demonstrating that Q-Cache enables effective multi-path exploration with high-quality information propagation. The complete framework with SR achieves the best performance across all models, with particularly strong gains for smaller LLMs (Qwen2.5-7B: 22.90\% to 56.44\%). These results validate that Q-Cache serves as the critical enabler for effective multi-path generation.

\begin{table*}[!t]
\centering
\footnotesize
\caption{Token cost comparison across methods on VerilogEval 2.0 and RTLLM 2.0 benchmarks. Token counts are averaged per problem. Multipliers ($\times$) are relative to Baseline.}
\label{tab:token_cost}
\resizebox{\textwidth}{!}{
\renewcommand{\arraystretch}{0.85}
\setlength{\aboverulesep}{2pt}
\setlength{\belowrulesep}{2pt}
\begin{tabular}{llcccccccc}
\toprule
\multirow{2}{*}{\textbf{Dataset}} & \multirow{2}{*}{\textbf{Model}} & \multicolumn{2}{c}{\textbf{Baseline}} & \multicolumn{2}{c}{\textbf{VeriMaAS}} & \multicolumn{2}{c}{\textbf{\ourtool-Lite}} & \multicolumn{2}{c}{\textbf{\ourtool}} \\
\cmidrule(lr){3-4} \cmidrule(lr){5-6} \cmidrule(lr){7-8} \cmidrule(lr){9-10}
& & \textit{Pass@1} & \textit{Tokens} & \textit{Pass@1} & \textit{Tokens} & \textit{Pass@1} & \textit{Tokens} & \textit{Pass@1} & \textit{Tokens} \\
\midrule
\multirow{2}{*}{VerilogEval 2.0} 
& Qwen2.5-Coder-32B & 46.93\% & 0.63k & 56.67\% & 3.67k (5.83$\times$) & 67.93\% & 3.73k (5.92$\times$) & 73.31\% & 6.95k (11.03$\times$) \\
& GPT-4o-mini & 48.97\% & 0.52k & 57.24\% & 3.18k (6.12$\times$) & 68.41\% & 2.94k (5.65$\times$) & 72.43\% & 5.51k (10.60$\times$) \\
\midrule
\multirow{2}{*}{RTLLM 2.0} 
& Qwen2.5-Coder-32B & 40.40\% & 0.76k & 55.82\% & 4.59k (6.04$\times$) & 62.82\% & 4.05k (5.33$\times$) & 65.49\% & 7.28k (9.58$\times$) \\
& GPT-4o-mini & 46.60\% & 0.67k & 57.25\% & 4.32k (6.45$\times$) & 61.29\% & 3.79k (5.66$\times$) & 64.23\% & 6.47k (9.66$\times$) \\
\bottomrule
\end{tabular}
}
\end{table*}

\textbf{Controlled Experiment for Multi-Path Benefits.} To scientifically validate whether the performance improvements of multi-path generation stem from inherent language properties rather than mere prompt variance, we conduct a controlled experiment. We compare two configurations using GPT-4o-mini with 4 layers: (1)~Two-Path (C++ + Python), using C++ and Python as intermediate representations; and (2)~Two-Path Python (bit-level + high-level), using two Python agents where one is prompted to generate bit-level style code and the other generates high-level abstract code. Two-Path (C++ + Python) achieves 64.52\% Pass@1 on VerilogEval 2.0 and 57.89\% on RTLLM 2.0, while Two-Path Python achieves 59.47\% and 53.22\% respectively. The consistent performance gaps of 5.05\% and 4.67\% demonstrate that benefits originate from inherent language properties: C++ provides native support for bit-level operations through built-in bitwise operators, explicit type control, and pointer manipulation, which cannot be fully replicated by prompting Python to mimic this style. This structural difference produces genuinely diverse intermediate representations, leading to more effective solution space exploration.

\begin{figure}[t]
  \centering
  \includegraphics[width=\columnwidth]{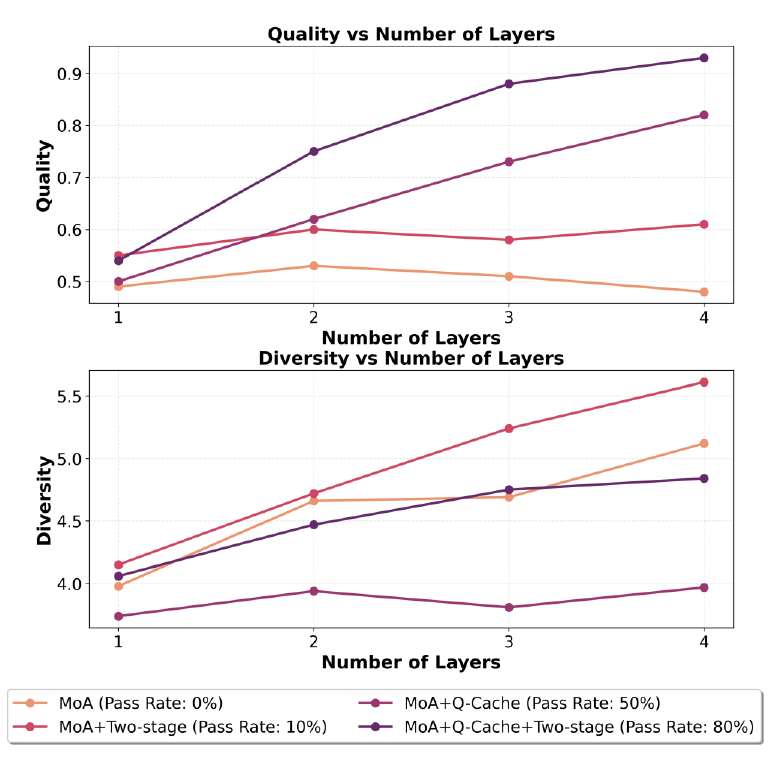}
  \caption{Quality and diversity metrics during LIFObuffer generation from RTLLM 2.0 across 10 trials.}
  \label{fig:quality_diversity}
\end{figure}

\subsection{Parameter Sensitivity Analysis (RQ3)}

We investigate the impact of two key hyperparameters on QMMoA's performance: layer number (depth) and layer width (number of parallel agents per layer). Figure~\ref{fig:heatmap} presents Pass@1 results on VerilogEval 2.0 using Qwen2.5-7B across different configurations.

The results reveal that both dimensions must reach sufficient scale for strong performance. With minimal width (1 agent) or shallow depth (1 layer), performance remains low (25.5\%-39.8\%). Significant improvements emerge only when both parameters reach adequate thresholds: configurations with 3+ layers and 4+ agents achieve 47.6\%-56.4\% Pass@1, substantially outperforming sparse configurations. This validates that our framework requires sufficient capacity in both dimensions to effectively accumulate high-quality solutions and explore diverse reasoning paths.

When comparing configurations with equal total agents, increasing width yields greater gains than increasing depth. For instance, with 12 total agents, the 3-layer-4-width configuration achieves 47.6\% versus the 4-layer-3-width configuration's 46.4\%. Furtherly,  2-layer-6-width achieves 48.5\%, outperforming the former both. This demonstrates that width (parallel diversity) maters importantly, validating our multi-path generation strategy's emphasis on exploring diverse solution spaces through heterogeneous agent mixtures within each layer.

\subsection{Inference Cost Analysis (RQ4)}
\label{sec:cost}

A key practical consideration for multi-agent frameworks is computational overhead. We conduct comprehensive cost experiments using GPT-4o-mini and Qwen2.5-Coder-32B on both benchmarks, comparing four methods: Baseline (direct generation), VeriMaAS (adaptive multi-agent workflow), \ourtool-Lite (our framework with 3 agents per layer), and \ourtool (our full framework with 6 agents per layer). Both \ourtool variants use 4 proposer layers with early stopping that terminates generation when a solution passes verification.

Table~\ref{tab:token_cost} presents the token cost comparison. \ourtool achieves +17.63\% to +26.38\% Pass@1 improvement over Baseline with 9.58$\times$ to 11.03$\times$ token consumption. Compared to VeriMaAS, \ourtool provides 2--3$\times$ larger performance gains while using only 1.7--1.8$\times$ more tokens. Under comparable token budgets ($\sim$5.5--6$\times$), \ourtool-Lite significantly outperforms VeriMaAS: on VerilogEval 2.0 with GPT-4o-mini, \ourtool-Lite achieves +19.44\% at 5.65$\times$ tokens, while VeriMaAS achieves +8.27\% at 6.12$\times$ tokens, demonstrating superior token efficiency of our quality-guided caching and multi-path generation.

Table~\ref{tab:tokens_accuracy} shows the relationship between token consumption and accuracy across different layer widths. A clear diminishing returns pattern emerges: expanding from 2 to 3 agents yields +3.89\% improvement on VerilogEval 2.0, while expanding from 5 to 6 agents produces only +0.94\%. Similarly, on RTLLM 2.0, the improvement drops from +3.40\% (2$\to$3) to +0.65\% (5$\to$6). This convergence trend provides practitioners with actionable guidance for cost-performance trade-off selection, as inference time and API costs scale linearly with token consumption.

For comparison, VeriRL requires over 30 GPU-hours on 8$\times$A100 for training, while \ourtool requires no training and only incurs $\sim$10$\times$ inference token overhead, making it advantageous for rapid deployment across different LLM backbones.

\subsection{Robustness and Failure Mode Analysis (RQ5)}
\label{sec:robustness}

\textbf{Performance Without Golden Testbenches.} \ourtool's quality-guided mechanism relies on simulation-based evaluation, raising the question of robustness when high-quality testbenches are unavailable. To evaluate this, we replace golden testbenches with LLM-generated ones during generation: the same GPT-4o-mini backbone generates testbenches directly from specifications via CoT prompting, while golden testbenches are retained only for final evaluation. As shown in Table~\ref{tab:testbench}, \ourtool experiences only 2.83\%--4.59\% degradation with LLM-generated testbenches, while still substantially outperforming VeriMaAS (57.24\% Pass@1 on VerilogEval 2.0). This robustness stems from two design aspects: testbenches are used only in quality evaluation and optional self-refinement (contributing merely $\sim$2\% improvement), and our quality evaluation algorithm employs testbenches only for coarse-grained binary classification, while subsequent pattern-based analysis ensures reliable ranking even with noisy feedback.

\textbf{Intermediate Language Failure Mode Analysis.} The two-stage process introduces a potential failure mode where semantic errors in intermediate C++/Python code may propagate to the final HDL. We analyze all intermediate code generated by GPT-4o-mini on RTLLM 2.0: 94.25\% is functionally correct, while only 5.75\% contains semantic errors. For correct intermediate code, 67.94\% of the final Verilog passes functional verification; when intermediate code is incorrect, only 3.48\% succeeds. The high Stage~1 correctness rate confirms that LLMs effectively leverage their high-level language fluency, and while error propagation exists, it occurs in only 5.75\% of cases, making this failure mode rare relative to the overall benefits of the two-stage process.

\begin{table}[t]
\centering
\footnotesize
\caption{Tokens vs.\ accuracy with varying agents per layer using GPT-4o-mini. Improvements and multipliers are relative to Baseline.}
\label{tab:tokens_accuracy}
\resizebox{\columnwidth}{!}{
\renewcommand{\arraystretch}{0.85}
\setlength{\aboverulesep}{2pt}
\setlength{\belowrulesep}{2pt}
\begin{tabular}{ccccc}
\toprule
\multirow{2}{*}{\textbf{Agents}} & \multicolumn{2}{c}{\textbf{VerilogEval 2.0}} & \multicolumn{2}{c}{\textbf{RTLLM 2.0}} \\
\cmidrule(lr){2-3} \cmidrule(lr){4-5}
& \textit{Pass@1} & \textit{Tokens} & \textit{Pass@1} & \textit{Tokens} \\
\midrule
2 & 64.52\% & 1.86k (3.57$\times$) & 57.89\% & 2.28k (3.40$\times$) \\
3 & 68.41\% & 2.94k (5.65$\times$) & 61.29\% & 3.79k (5.66$\times$) \\
4 & 70.28\% & 3.88k (7.46$\times$) & 62.76\% & 4.71k (7.03$\times$) \\
5 & 71.49\% & 4.70k (9.04$\times$) & 63.58\% & 5.60k (8.36$\times$) \\
6 & 72.43\% & 5.51k (10.60$\times$) & 64.23\% & 6.47k (9.66$\times$) \\
\bottomrule
\end{tabular}
}
\end{table}

\begin{table}[t]
\centering
\footnotesize
\caption{\ourtool performance with golden vs.\ LLM-generated testbenches using GPT-4o-mini.}
\label{tab:testbench}
\resizebox{\columnwidth}{!}{
\renewcommand{\arraystretch}{0.85}
\setlength{\aboverulesep}{2pt}
\setlength{\belowrulesep}{2pt}
\begin{tabular}{lcccccc}
\toprule
\multirow{2}{*}{\textbf{Testbench}} & \multicolumn{3}{c}{\textbf{VerilogEval 2.0 (\%)}} & \multicolumn{3}{c}{\textbf{RTLLM 2.0 (\%)}} \\
\cmidrule(lr){2-4} \cmidrule(lr){5-7}
& \textit{P@1} & \textit{P@3} & \textit{P@5} & \textit{P@1} & \textit{P@3} & \textit{P@5} \\
\midrule
Golden & 72.43 & 76.94 & 79.46 & 64.23 & 67.45 & 68.67 \\
LLM-Gen. & 67.84 & 73.57 & 76.63 & 60.51 & 63.87 & 65.74 \\
\midrule
$\Delta$ & \textit{-4.59} & \textit{-3.37} & \textit{-2.83} & \textit{-3.72} & \textit{-3.58} & \textit{-2.93} \\
\bottomrule
\end{tabular}
}
\end{table}

\subsection{Case Study: Quality and Diversity Analysis (RQ6)}
\label{sec:case_study}

We analyze quality and diversity evolution on a complex task from RTLLM 2.0, named "LIFObuffer", by using Qwen2.5-7B with layer width of 6. We measure the top-6 cached HDL candidates at each layer across four configurations, averaging metrics over 10 trials. Quality is evaluated using our scoring mechanism, while diversity is measured by Vendi score~\cite{li2025rethinking}. Figure~\ref{fig:quality_diversity} reveals distinct patterns.
Standard MoA and MoA+Two-stage show stagnant quality (0.49 to 0.48 and 0.55 to 0.61) with 0-10\% pass rates. MoA+Q-Cache achieves monotonic quality improvement (0.50 to 0.82, 50\% pass rate), while MoA+Q-Cache+Two-stage demonstrates the strongest growth (0.54 to 0.93, 80\% pass rate). This validates that Q-Cache enables progressive quality improvement, which Two-stage amplifies when built upon this stable foundation.

For diversity, MoA and MoA+Two-stage maintain high values (3.98-5.61) but achieve low pass rates, indicating unfocused exploration. MoA+Q-Cache shows modest diversity (3.74-3.97) but reasonable performance through quality filtering. MoA+Q-Cache+Two-stage achieves both high diversity (4.06-4.84) and the best pass rate (80\%), confirming that diversity alone is insufficient—effective exploration requires quality-guided selection to ensure diverse candidates contribute meaningful improvements rather than noise.
\section{Conclusion and Future Work}

This paper introduces \ourtool, a quality-guided multi-path Mixture-of-Agents framework that systematically addresses noise susceptibility and constrained reasoning space in multi-agent HDL generation. By combining a quality-guided caching mechanism that ensures monotonic knowledge accumulation with a multi-path generation strategy leveraging C++ and Python as intermediate representations, \ourtool achieves 15--30\% improvements in Pass@1 across diverse LLM backbones on both VerilogEval 2.0 and RTLLM 2.0 benchmarks. Our framework enables smaller LLMs to match and even exceed larger models and fine-tuned alternatives without costly training. Furthermore, \ourtool provides complementary benefits when applied on top of fine-tuned models, achieving +16.19\% improvement over VeriRL-CodeQwen2.5 on VerilogEval 2.0, demonstrating that effective architectural design through quality-guided progress and diverse solution exploration offers a practical and scalable approach for automated RTL design.

\textbf{Limitations and Future Work.} While VerilogEval 2.0 and RTLLM 2.0 are widely adopted benchmarks ensuring fair comparison, they consist of relatively small-scale designs. For complex real-world designs such as RISC-V cores or NoC routers, the context overhead of carrying multiple intermediate representations and cached candidates may pose challenges. A practical direction is hierarchical generation: a planner agent decomposes the overall design specification into manageable module tasks, \ourtool generates each module independently, and the modules are integrated to form the complete design. This aligns with standard hardware design methodology where complex systems are naturally structured as hierarchical modules.

% Acknowledgements should only appear in the accepted version.
\section*{Acknowledgements}
The authors H.P., P.Z., S.L., A.C., X.Z., and P.B. acknowledge the support by the National Science Foundation (NSF) under the NSF Award 2243104 under the Center for Complex Particle Systems (COMPASS), the NSF Mid-Career Advancement Award BCS-2527046, the U.S. Army Research Office (ARO) under Grant No. W911NF-23-1-0111, the National Institute of Health (NIH) R01 AG 079957 "Interpretable machine learning to synergize brain age estimation and neuroimaging genetics", the Defense Advanced Research Projects Agency (DARPA) Young Faculty Award and DARPA Director Fellowship Award under Grant Number N66001-17-1-4044, Intel faculty awards, Northrop Grumman grant, and the NIH grants R01 AG 079957  and RF1 AG 082201 "Neurovascular calcification and ADRD in two nonindustrial Native American populations". It was a wonderful experience designing and writing the grant application entitled "Neurovascular calcification and ADRD in two nonindustrial Native American populations" and awarded under RF1 AG 082201. The views, opinions, and/or findings in this article are those of the authors and should not be interpreted as official views or policies of the Department of War, the National Institute of Health or the National Science Foundation.

% \textbf{Do not} include acknowledgements in the initial version of
% the paper submitted for blind review.

% If a paper is accepted, the final camera-ready version can (and
% probably should) include acknowledgements. In this case, please
% place such acknowledgements in an unnumbered section at the
% end of the paper. Typically, this will include thanks to reviewers
% who gave useful comments, to colleagues who contributed to the ideas,
% and to funding agencies and corporate sponsors that provided financial
% support.

% In the unusual situation where you want a paper to appear in the
% references without citing it in the main text, use \nocite
% \nocite{langley00}

\bibliography{example_paper}
\bibliographystyle{mlsys2025}

%%%%%%%%%%%%%%%%%%%%%%%%%%%%%%%%%%%%%%%%%%%%%%%%%%%%%%%%%%%%%%%%%%%%%%%%%%%%%%%
%%%%%%%%%%%%%%%%%%%%%%%%%%%%%%%%%%%%%%%%%%%%%%%%%%%%%%%%%%%%%%%%%%%%%%%%%%%%%%%
% SUPPLEMENTAL CONTENT AS APPENDIX AFTER REFERENCES
%%%%%%%%%%%%%%%%%%%%%%%%%%%%%%%%%%%%%%%%%%%%%%%%%%%%%%%%%%%%%%%%%%%%%%%%%%%%%%%
%%%%%%%%%%%%%%%%%%%%%%%%%%%%%%%%%%%%%%%%%%%%%%%%%%%%%%%%%%%%%%%%%%%%%%%%%%%%%%%

% \appendix
% \section{Please add supplemental material as appendix here}
% %
% Put anything that you might normally include after the references as an appendix here, {\it not in a separate supplementary file}. Upload your final camera-ready as a single pdf, including all appendices.

%%%%%%%%%%%%%%%%%%%%%%%%%%%%%%%%%%%%%%%%%%%%%%%%%%%%%%%%%%%%%%%%%%%%%%%%%%%%%%%
%%%%%%%%%%%%%%%%%%%%%%%%%%%%%%%%%%%%%%%%%%%%%%%%%%%%%%%%%%%%%%%%%%%%%%%%%%%%%%%

\end{document}